\documentclass[10pt,twocolumn,letterpaper]{article}

\usepackage[pagenumbers]{cvpr} 

\definecolor{cvprblue}{rgb}{0.21,0.49,0.74}
\usepackage[pagebackref,breaklinks,colorlinks,allcolors=cvprblue]{hyperref}

\usepackage[utf8]{inputenc} 
\usepackage[T1]{fontenc}    
\usepackage{url}            
\usepackage{booktabs}       
\usepackage{amsfonts}       
\usepackage{nicefrac}       
\usepackage{microtype}      
\usepackage{xcolor}         
\usepackage{amsmath}
\usepackage{amssymb}
\usepackage{wrapfig}  
\usepackage{algorithm}
\usepackage{algorithmic}
\usepackage{multirow}
\usepackage{graphicx}
\usepackage{makecell} 

\def\paperID{*****} 
\def\confName{CVPR}
\def\confYear{2026}

\title{PathRelax: Parallel-Path Relaxed Speculative Jacobi Decoding for Accelerating Auto-Regressive Text-to-Image Generation}

\author{Haodong Lei\\
College of Software Engineering\\
Southeast University\\
{\tt\small leihaodong@seu.edu.cn}
\and
Hongsong Wang\\
School of Computer Science and Engineering\\
Southeast University\\
{\tt\small hongsongwang@seu.edu.cn}
\and
Bingxuan Dai\\
School of Cyber Science and Engineering\\
Southeast University\\
{\tt\small 220245799@seu.edu.cn}
\and
Pan Zhou\\
School of Computing and Information Systems\\
Singapore Management University\\
{\tt\small  panzhou@smu.edu.sg}
}

\begin{document}
\maketitle

\begin{abstract}
The growing need for high-resolution image generation in autoregressive text-to-image models has resulted in extended token sequences, significantly increasing computational costs and inference times. However, existing state-of-the-art methods for accelerating autoregressive text-to-image models rely on chain-structured draft token sequences, leading to inefficient draft token search and limited acceptance lengths. To address this, we propose parallel-path cross-relaxed speculative Jacobi decoding (\textbf{PathSpec}), a novel framework that enhances efficiency through a multi-sequence draft tree structure. Our parallel-path speculative Jacobi decoding (\textbf{PathExplore}) expands the token search space, achieving a higher speedup ratio without sacrificing image quality. Additionally, we introduce cross-path relaxed verification (\textbf{PathRelax}) that exploits semantic similarities across sequences to further boost token acceptance rates. Evaluated on the Parti-Prompts, MSCOCO2017, and T2ICompBench datasets, our method achieves a speedup ratio of 4.14×, 3.95×, and 4.18×, respectively. Remarkably, PathExplore, without any relaxed sampling, outperforms relaxed sampling methods in the speedup ratio, such as GSD and LANTERN. Moreover, PathRelax’s relaxation mechanism can be seamlessly integrated with other relaxation techniques, enabling further acceleration and providing an efficient solution for real-time text-to-image generation. Our code is available at \url{https://github.com/Haodong-Lei-Ray/PathSpec}.
\end{abstract}

\begin{figure*}[t]
    \centering
    \includegraphics[width=1\textwidth]    {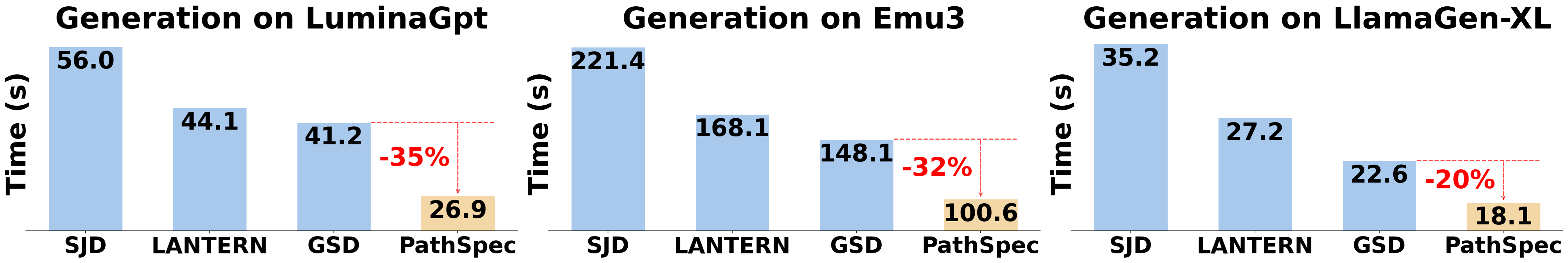}
    \caption{\textbf{Time cost comparison of PathSpec and other methods.} PathSpec achieves at least a 20\% reduction in image generation time across various models.}
    \label{fig:1.1}
    \vspace{-1em}
\end{figure*}
\section{Introduction}
\label{sec:intro}

Text-to-image generative models have achieved remarkable success in producing high-resolution photorealistic images for applications in gaming, filmmaking, and digital content creation~\cite{stable-diffusion,dall-e3,imagen}. However, this success comes at the cost of high inference latency—particularly for autoregressive (AR) models, which are increasingly favored for their flexibility and potential to unify multimodal generation within a single framework~\cite{chameleon,gemma2, EMU3,luminamgpt}. Unlike diffusion models~\cite{DDPM} that synthesize images in a low-dimensional latent space, AR models, e.g., Lumina-mGPT~\cite{luminamgpt} and Emu3~\cite{EMU3},  generate discrete image tokens sequentially, often requiring thousands of tokens per image, far exceeding typical text sequence lengths. This results in high inference latency, making real-time generation impractical. 

To mitigate inefficiency caused by the token-by-token mechanism, speculative decoding~\cite{SSM, SSM1}, which is originally designed for text LLMs, has been extended to autoregressive image models\cite{luminamgpt2, GSD}. Its key idea is to let a lightweight draft model or an internal mechanism generate multiple draft tokens, which are then verified in a single forward pass by the target image model. This strategy enables the generation of several valid tokens per step, and substantially accelerates inference, e.g., achieves a 1.88× speedup with Speculative Jacobi Decoding (SJD) on Lumina-mGPT~\cite{luminamgpt}. Such advances highlight speculative decoding as a promising direction for efficient large-scale image generation. 

Existing speculative decoding methods for image generation fall into two categories. Cross-domain adaptation approaches, such as LANTERN~\cite{LANTERN} and LANTERN++~\cite{LANTERN++}, transfer text-based speculative decoding to vision via relaxed sampling (often below 48\% acceptance~\cite{LANTERN}), but rely on separately trained draft models that may underfit visual token distributions. Self-speculative methods, including SJD~\cite{SJD} and GSD~\cite{GSD}, remove the draft model to avoid additional training, yet remain constrained by chain-structured dependencies, where rejection of an early token invalidates subsequent drafts, limiting acceptance length and acceleration ($\approx$ 1.88 accepted tokens per 16 draft tokens of SJD on Lumina-mGPT). Moreover, existing relaxation strategies~\cite{GSD,LANTERN++} operate at single-position token similarity in the codebook space~\cite{VQVAE}, overlooking structural dependencies across draft tree branches.

To address these limitations, we introduce Parallel-Path Cross-Relaxed Speculative Jacobi Decoding (\textbf{PathSpec}), a new speculative sampling paradigm that expands both the search structure and relaxation space for image token generation, thereby improving token acceptance rates and decoding efficiency while preserving distributional consistency.

To alleviate the acceptance length limitation caused by chain-structured dependency in conventional self-speculative decoding frameworks, we design Parallel-Path Speculative Jacobi Decoding (\textbf{PathExplore}). It generalizes the linear draft chain into a SpecInfer-inspired multi-sequence tree, where multiple candidate token sequences are expanded in parallel and jointly verified by the target model. As a result, rejection of one branch does not invalidate others, mitigating the cascading discard issue inherent in chain-based decoding. Even without relaxation, PathExplore substantially improves the average acceptance length while preserving visual fidelity, achieving a 3.06× gain over SJD.

Furthermore, we propose cross-path relaxed verification (\textbf{PathRelax}) to address the limitation of existing approaches that rely solely on single-token similarity for local relaxation while ignoring semantic dependencies across branches. In multi-sequence tree structures, different paths often share similar semantic prefixes and differ only in local tokens. To leverage this property, we introduce cross-path probability aggregation to establish associations among semantically related branch positions, and employ acceptance with relaxed verification to enable path-level information sharing, such that the acceptance of one branch can facilitate the acceptance of its semantic neighbors. The proposed strategy enlarges the relaxation space and significantly enhances decoding stability and acceleration efficiency.

Together, these innovations enable our method to significantly improve token acceptance length and decoding efficiency in text-to-image generation while preserving visual quality. As shown in Fig.~\ref{fig:1.1}, PathSpec consistently reduces image generation time across different models, achieving at least a 20\% improvement in efficiency. This improvement alleviates inference overhead while preserving generation quality~\cite{clipscore,hpsv2}, enhancing scalability for real-world applications~\cite{pixart-alpha}. 

\section{Related Work}
\label{sec:relatedwork}

\noindent \textbf{Visual Autoregressive Models:} AR models have gained prominence in image generation, delivering quality rivaling diffusion models~\cite{LDM-4} through sequential token prediction. Unlike diffusion models, visual AR models tokenize images into discrete sequences~\cite{ViT} and process them with transformer architectures, the same to large language models (LLMs). Existing works such as LlamaGen~\cite{Llamagen}, Anole~\cite{Anole}, Emu3~\cite{EMU3}, and Lumina-mGPT~\cite{luminamgpt, luminamgpt2} excel in text-conditional image generation, using quantized autoencoders to convert images into token sequences for transformer-based sampling.
\\
\\
\noindent \textbf{Speculative Decoding:}
The core idea of speculative decoding \cite{SSM, SSM1} is to first draft and then verify: quickly generate a potentially correct draft and then check which tokens in the draft can be accepted. This method first applies to large language models with an autoregressive structure. The initial draft form is the chain structure~\cite{JD,lookhead}. And then SpecInfer \cite{SpecInfer} introduces a draft form with a tree structure, which represents a draft tree. The draft form with tree structure~\cite{SpecInfer, MEDUSA, EAGLE-2, HASS} has flourished. From MEDUSA~\cite{MEDUSA} to EAGLE-2~\cite{EAGLE-2}, unleashing the potential of the tree structure draft tree, these methods greatly increase the speed-up ratio in large language models.

One of the few works related to speculative decoding of image token sequences is speculative decoding for Multi-LLM \cite{SPD4MLLM, SpecVLM, Spec-LLaVA, SpecPrune-VLA}, which provides an efficient approach to applying speculative decoding in the visual understanding model. With the introduction of SJD~\cite{SJD}, speculative decoding is extended to visual generation AR models. LANTERN~\cite{LANTERN++, LANTERN} applied relaxed sampling to group similar codebook tokens in EAGLE-2~\cite{EAGLE-2}, but is limited by low draft model accuracy. GSD~\cite{GSD} extended relaxed sampling to SJD, but its speedup is restricted by SJD’s single-sequence design.
COOL-SD~\cite{coolsd} boosts acceleration by employing a relaxed strategy based on an annealed acceptance schedule, while VVS~\cite{VVS} improves efficiency by partially skipping the verification stage and reusing intermediate features. In addition, recent work has explored introducing more sophisticated processing within the draft phase. For example, MC-SJD~\cite{MC-SJD} and SJD++~\cite{SJD++} investigate how preserving high-value draft tokens can further enhance acceleration. SJD-PAC~\cite{SJD-PAC} extends the accepted sequence length through an adaptive continuation mechanism. Meanwhile, a growing line of research integrates diffusion architectures with speculative sampling~\cite{DiffSpec, ding2026inductive}. Along this direction, SJD2~\cite{SJD2} explores a new paradigm that unifies diffusion models with AR models.

\section{Preliminaries}
\noindent\textbf{Autoregressive Text-to-Image Generation:}
Modern text-to-image models generate images by autoregressively predicting discrete visual tokens in a learned latent space (e.g., via VQ-based tokenization).
Given text tokens as context, an autoregressive model samples image tokens sequentially according to:
\begin{equation}
	p_\theta(\mathbf{x}_{1:L}) = \prod_{i=1}^{L} p_\theta(\mathbf{x}_i \mid \mathbf{x}_{1:i-1}),
\end{equation}
where one token is produced per forward pass.
While effective, this strictly sequential decoding results in high inference latency for high-resolution image generation, motivating parallelizable acceleration strategies.
\\
\\
\noindent\textbf{Speculative Decoding:}
It is a lossless inference technique designed to accelerate autoregressive generation without altering the target distribution $p_\theta$. The key idea is to decouple the token proposal from the token verification.
Instead of sampling a single token at each step, speculative decoding first generates several candidate tokens (``draft tokens'') from a faster draft distribution $q$, and then verifies them in parallel using the target model $p_\theta$.  
Concretely, for a draft token $\mathbf{x}$, acceptance is determined by
\begin{equation}\label{acceptingrule}
r_i \leq	\min\!\left(1, \frac{p_\theta(\mathbf{x} \mid \mathbf{x}_{1:i-1})}{q(\mathbf{x} \mid \mathbf{x}_{1:i-1})}\right),
\end{equation}
where $r_i$ is sampled from a uniform distribution, i.e., $r_i \sim \mathcal{U}[0,1]$. This guarantees that accepted tokens are exact samples from $p_\theta$.
When $q$ is cheap and well-aligned with $p_\theta$, multiple draft tokens are accepted per iteration, advancing several AR steps per forward pass and yielding substantial speedups.
\\
\\
\noindent\textbf{Speculative Jacobi Decoding (SJD):}
\label{sec:Pre-SJD}
It reuses the target model’s own predictions from previous iterations as draft tokens, enabling efficient construction of draft distribution $p$. Formally, at decoding iteration $j$, SJD updates a contiguous block of $W$ tokens (the Jacobi window) in parallel.
Let $\mathbf{x}_i^{(j)}$ denote the token at position $i$ in iteration $j$.
The draft distribution for token $\mathbf{x}_i^{(j)}$ is defined as the target model’s prediction from the previous iteration:
\begin{equation}
	q\!\left(\mathbf{x}_i^{(j)} \mid \mathbf{x}_{1:i-1}\right)
	\;=\;
	p_\theta\!\left(\mathbf{x}_i^{(j)} \mid \mathbf{x}_{1:i-1}^{(j-1)}\right).
\end{equation}

Each draft token is then independently verified using the current iteration’s token sequence.
Acceptance follows the standard speculative rule as shown in Eq.~\eqref{acceptingrule}.
By construction, SJD preserves exact sampling from $p_\theta$ while enabling multiple tokens to be generated and verified in parallel at each iteration.
This Jacobi-style update significantly reduces the number of sequential forward passes required for autoregressive text-to-image generation, making SJD a scalable acceleration framework.

\begin{figure*}[t]
	\centering
	\includegraphics[width=1\linewidth]{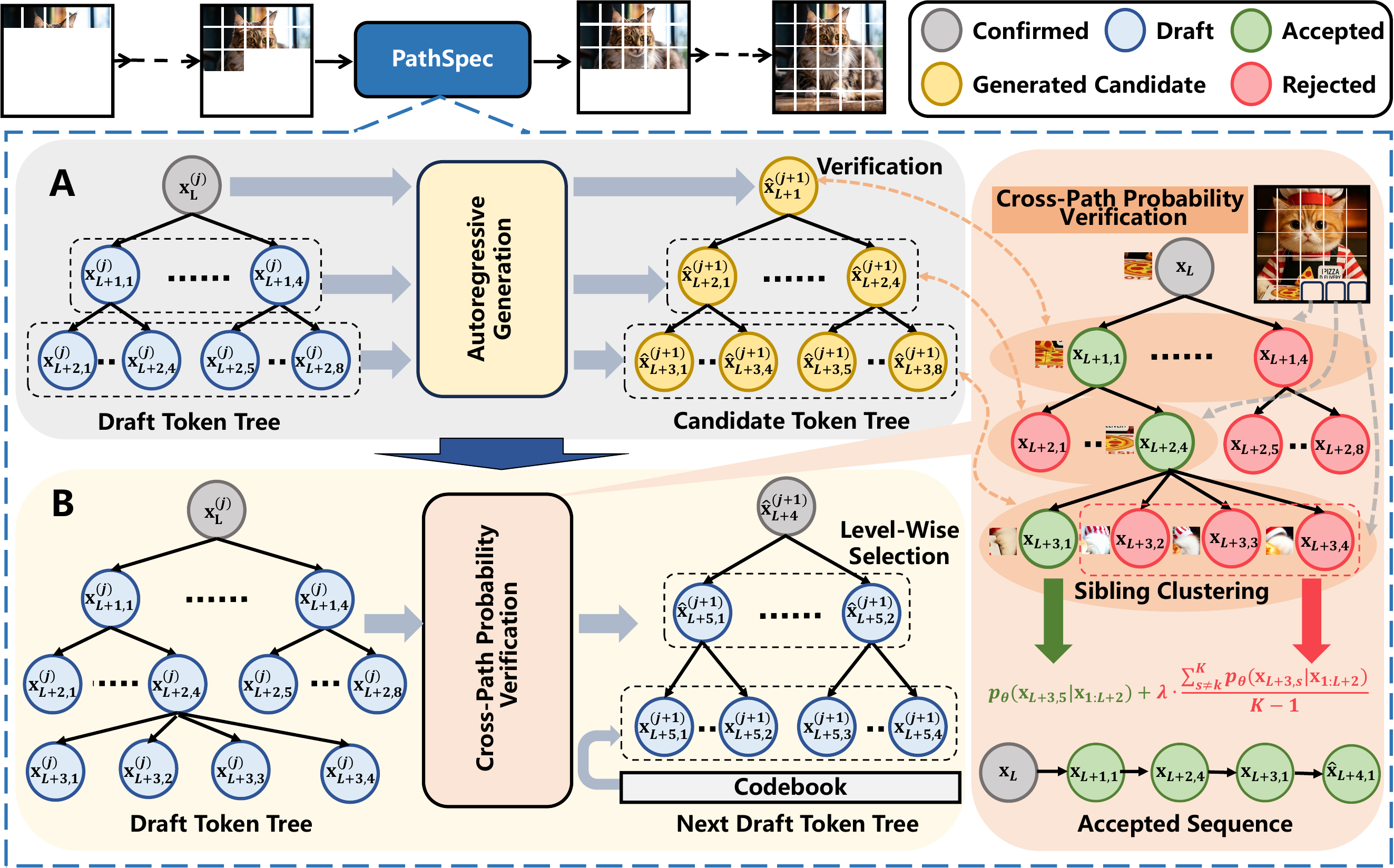}
	\caption{\textbf{Overview of the proposed PathSpec.} As shown in part A of the figure, this illustrates the target model's decoding process over the tree-structured draft tokens.
    Part B of the figure illustrates the \emph{Cross-Path Relaxed Verification} process applied to the Candidate Token Tree and the Draft Token Tree.
    }
	\label{fig:main}
    \vspace{-1em}
\end{figure*}
\section{Methodology}
\label{sec:method}
\textbf{PathSpec} is a tree-structured parallel-path cross-relaxed speculative decoding framework for autoregressive text-to-image generation that enables structured multi-branch exploration and cross-path relaxed verification, facilitating semantic-aware probability aggregation and acceptance sharing across branches, as illustrated in Fig.~\ref{fig:main}.
PathSpec consists of two key components: 
(i) \textbf{PathExplore}, which expands multiple speculative paths in parallel to increase acceptance length; and
(ii) \textbf{PathRelax}, which relaxes verification by sharing probability mass across aligned paths.

\subsection{Parallel-Path Speculative Jacobi Decoding}
\label{sec:multi-seq-sjd}
This section provides a detailed description of PathExplore. Existing speculative Jacobi decoding approaches often introduce tree-structured candidate spaces to improve acceptance rates, but they typically rely on a separate draft model to generate the draft tree at each iteration. Such dependence can limit robustness when the draft model is poorly aligned with the target model. In contrast, PathExplore does not assume an external draft generator. Instead, it initializes the draft tree via random token sampling from the codebook and generates candidate nodes through autoregressive next-token prediction conditioned on their prefix sequences. The constructed tree is then verified in a single parallel pass against the target model outputs, forming an efficient multi-branch exploration mechanism without requiring a dedicated draft model.

Specifically, let $\mathcal{T}^{(j)}$ denote the draft token tree maintained at the $j$-th speculative Jacobi iteration. Unlike conventional speculative decoding, our method constructs and updates the draft tree directly from token-level operations. At initialization, draft tokens are randomly sampled from the discrete codebook to populate the tree structure. During each iteration, speculative verification is performed along all branches. Let the verified prefix sequence be denoted as $\mathbf{x}^{(j)}_{1:L}$, where $\mathbf{x}^{(j)}_L$ is the position of the last accepted token and becomes the root node of the next draft tree. We use any representative path of the draft token tree in Fig.\ref{fig:main}(A) to describe the proposed strategy, which is represented as:
\begin{equation}
\mathbf{x}^{(j)}_{L:L+d}=(\mathbf{x}^{(j)}_L,\mathbf{x}^{(j)}_{L+1},\dots,\mathbf{x}^{(j)}_{L+d}),
\end{equation}
where $d$ represents the depth of the tree.
The subsequent draft token is sampled from the conditional distribution defined by the target model:
\begin{equation}
\left\{\hat{\mathbf{x}}^{(j+1)}_{L+i+1} \right\}_{i=0}^{d}
\sim p_\theta\!\left(\cdot \mid \mathbf{x}^{(j)}_{1:L+i}\right),
\end{equation}
where $p_\theta(\cdot \mid \cdot)$ denotes the autoregressive prediction function with model parameter $\theta$. All candidate nodes within the same tree level are generated in parallel using a shared target-model forward pass with cached prefix states. The generated candidate tokens are aligned with the corresponding nodes in the constructed draft tree and evaluated using the previous verification method\cite{SSM} to determine their retention.

\begin{figure*}[t]
	\centering
	\begin{minipage}[t]{0.45\textwidth}
		\centering
		\includegraphics[width=\textwidth, height=0.15\textheight,keepaspectratio=false]{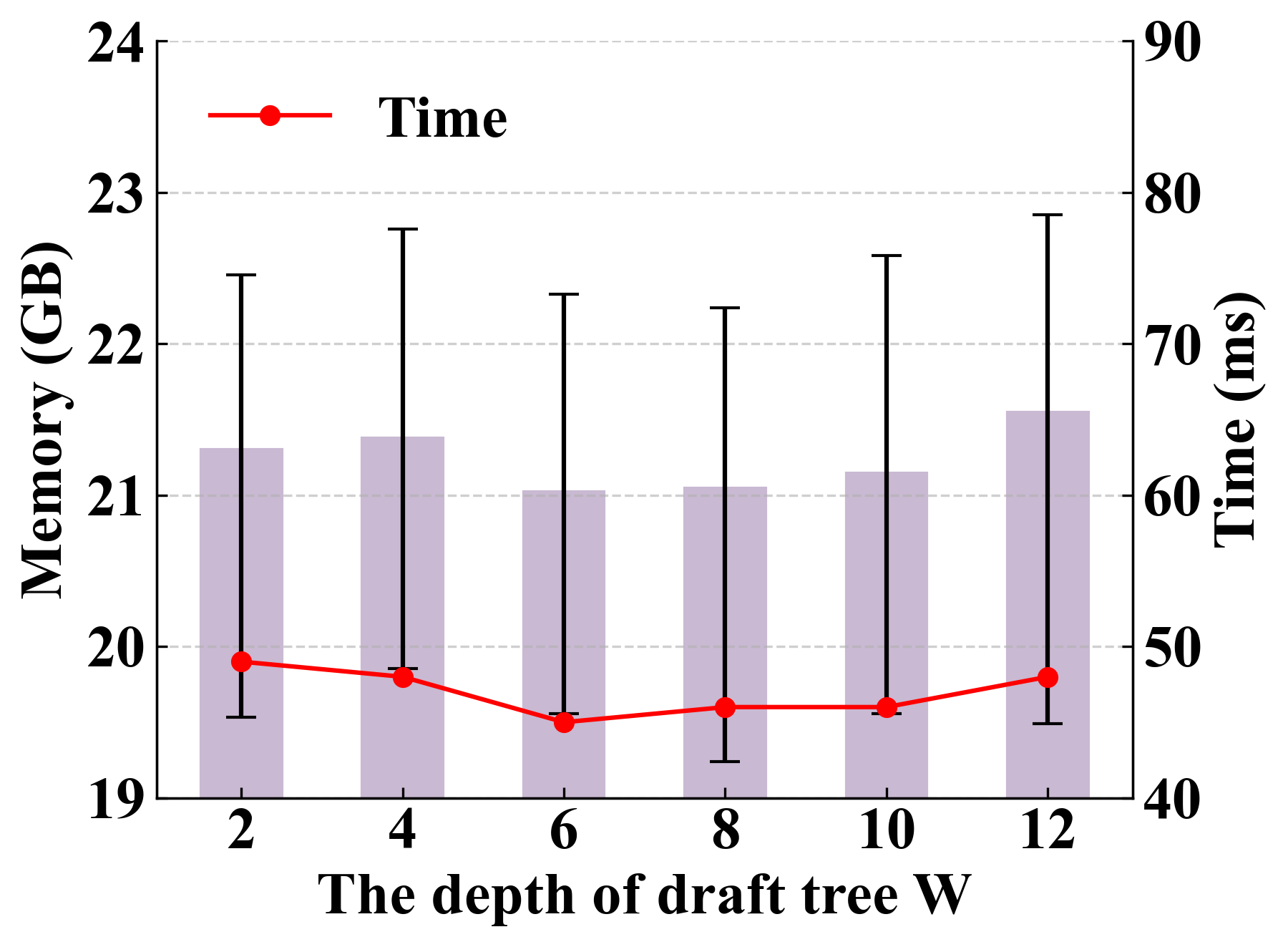}
		\subcaption{}
	\end{minipage}
	\begin{minipage}[t]{0.45\textwidth}
		\centering
		\includegraphics[width=\textwidth, height=0.15\textheight,keepaspectratio=false]{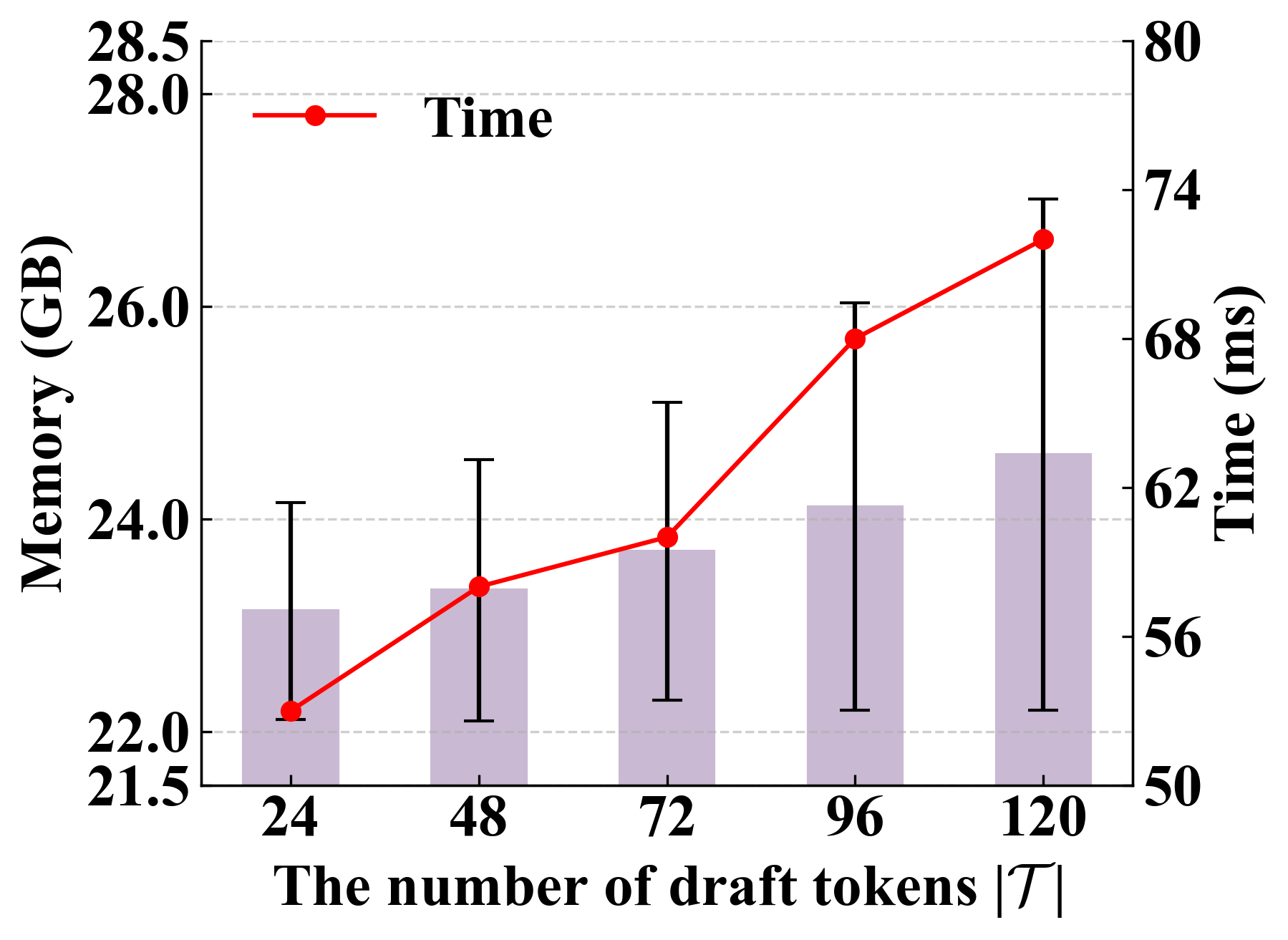}
		\subcaption{}
	\end{minipage}
	\caption{
    \textbf{The comparison of the cost (time and memory) of PathExplore for Lumina-GPT-7B model.}
    (a) The memory usage and one forward time variations across different $W$ values.
    (b) The memory usage and one forward time variations across different $|\mathcal{T}|$ values.
    }
    \label{fig:time_space}
    \vspace{-1em}
\end{figure*}

After verification of $\mathcal{T}^{(j)}$, the retained candidate tokens are assigned to different levels of $\mathcal{T}^{(j+1)}$ according to their positional encodings, as shown in Fig.~\ref{fig:main}(B). Due to node capacity constraints, we adopt a \textbf{tree-position priority selection} strategy, where tokens with smaller $z$ values are preferentially retained. Unassigned positions are randomly initialized from the image vocabulary. The KV cache of accepted prefixes is reused across iterations, while sibling branches inherit cached states from shared parent nodes. This enables PathExplore to reuse draft tokens from $\mathcal{T}^{(j)}$ when constructing $\mathcal{T}^{(j+1)}$, reducing generation overhead and improving token acceptance during verification.

Now we discuss how PathExplore can increase the number of accepted draft tokens. Assume that the node corresponding to token $\mathbf{x}^{(j)}_{i-1}$ in the draft tree $\mathcal{T}^{(j)}$ has $K$ child nodes, associated with candidate draft tokens $\mathbf{x}^{(j)}_{i,1}, \mathbf{x}^{(j)}_{i,2}, \dots, \mathbf{x}^{(j)}_{i,K}$, respectively, where $i$ is the position where the current token was generated. Since a token position is accepted if at least one candidate path succeeds, the effective acceptance probability $\alpha_i$ can be written as:
\begin{equation}
	\alpha_i
	=
	1 - \prod_{k=1}^{K} \left(1 - \alpha_{i,k}\right)
	\;\ge\;
	\max_k \alpha_{i,k}.
\end{equation}
This multi-path exploration mechanism increases the expected token acceptance rate compared with single-path speculative decoding, thereby reducing the number of decoding iterations.
Finally, we analyze the inference cost of PathExplore, including the cost of creating the draft tree and the verification cost: (1) For generation cost, PathExplore follows a self-speculative decoding paradigm\cite{SJD,SJD++} and does not rely on auxiliary draft models. Draft tokens are initialized using the ISP\cite{SJD} method or sampled autoregressively from the token at position $i-1$ in the $j$-th iteration. Therefore, draft-token generation is implemented through a small number of parallelized target-model forward passes, requiring approximately one autoregressive forward pass to generate $|\mathcal{T}|$ draft tokens with negligible additional latency. Each draft token is stored as an integer index (typically 4 bytes), resulting in only a few hundred bytes of memory overhead under practical settings. (2) For verification cost, PathExplore adopts a tree-mask mechanism~\cite{SpecInfer} to avoid linear verification complexity in $|\mathcal{T}|$, enabling simultaneous validation of all draft tree sequences within a single forward pass of the target model. Benefiting from the parallel computation capability of Transformer architectures, tree-path verification introduces only modest additional computational overhead.

This introduces only modest memory and computational overhead. As shown in Fig.~\ref{fig:time_space}(a), increasing tree depth with fixed width ($B=4$) results in no substantial growth in memory or forward time. Similarly, Fig.~\ref{fig:time_space}(b) shows that when the total number of nodes is kept below 72, the overhead remains well controlled. As a result, inference latency is comparable to SJD, while the acceptance length increases significantly. By selecting the sequence with the longest accepted prefix at each iteration, PathExplore achieves an effective space--time trade-off, reducing autoregressive depth with minimal memory cost.

\subsection{Cross-Path Relaxed Verification}
\label{sec:cross-relaxed}

In contrast, PathExplore exposes a new and previously unexplored opportunity: Under the same parent node, tokens generated at the same position across different speculative paths are frequently semantically aligned and probabilistically similar. This multi-path structure enables a fundamentally different relaxation mechanism—\emph{across paths rather than within a sequence}. We leverage this insight to propose PathRelax, a novel relaxation strategy that requires no auxiliary models, no codebook-specific assumptions, and integrates naturally into speculative rejection sampling.

\noindent\textbf{Cross-Path Probability Aggregation:}
As illustrated in the \textit{Cross-Path Probability Aggregation} module in Fig.~\ref{fig:main}, probability aggregation is performed locally within each sibling group, that is, only among child nodes that share the same parent prefix, rather than across all nodes at the same depth of the tree. Consider a parent node corresponding to the shared prefix $\mathbf{x}^{(j)}_{1:L-1}$. From this prefix, the draft model expands $K$ candidate child nodes at level $L$, denoted as $\{\mathbf{x}^{(j)}_{L,k}\}_{k=1}^{K}$. In conventional tree decoding, each child branch is independently evaluated using its draft model probability $q_\theta(\mathbf{x}^{(j)}_{L,k} \mid \mathbf{x}^{(j-1)}_{1:L-1})$. However, we first perform a sibling clustering operation, grouping together all children originating from the same parent node. Within each sibling group, we introduce cross-path probability aggregation to encourage semantically consistent alternatives under the same parent to reinforce each other. Specifically, for each child branch $\mathbf{x}^{(j)}_{L,k}$, we define the aggregated target probability as:
\begin{equation}
	\begin{split}
		& p_{\text{target}}\!\left(\mathbf{x}^{(j)}_{L,k} \mid \mathbf{x}^{(j)}_{1:L-1}\right)
        \\& = 
		\;\underbrace{
			p_\theta\!\left(\mathbf{x}^{(j)}_{L,k} \mid \mathbf{x}^{(j)}_{1:L-1}\right)
		}_{\text{self-path}}
		+
		\underbrace{
			\lambda \cdot \frac{\sum_{s \neq k}^K p_\theta\!\left(\mathbf{x}^{(j)}_{L,s} \mid \mathbf{x}^{(j)}_{1:L-1}\right)}{K-1}
		}_{\text{cross-path}},
	\end{split}
\end{equation}
where $K$ is the number of child nodes of $\mathbf{x}^{(j)}_{L-1}$ and $\lambda$ is the cross-path correlation factor which controls the strength of relaxation.
The first term corresponds to the intrinsic draft probability along the current branch (self-path), while the second term aggregates probability mass from its sibling branches (cross-path). The normalization factor $1/(K-1)$ prevents the aggregated contribution from scaling with the number of siblings, and the hyperparameter $\lambda$ controls the strength of cross-branch interaction. When $\lambda=0$, the formulation reduces to standard independent tree expansion. As $\lambda$ increases, branches under the same parent become more cooperative, effectively smoothing local competition while preserving inter-parent independence.

\noindent\textbf{Acceptance with Relaxed Verification:}
Speculative verification proceeds via rejection sampling using the relaxed target.
A draft token $\mathbf{x}_{L,k}$ is accepted if
\begin{equation}
	\min\!\left(
	1,
	\frac{
		p_{\text{target}}\left(\mathbf{x}^{(j)}_{L,k} \mid \mathbf{x}_{1:L-1}^{(j)}\right)
	}{
		p_\theta\!\left(\mathbf{x}^{(j)}_{L,k} \mid \mathbf{x}_{1:L-1}^{(j-1)}\right)
	}
	\right)
	>
	r,
	\quad r \sim \mathcal{U}[0,1],
\end{equation}
where $p_{\text{target}}$ denotes the probability obtained from the \textit{cross-path probability aggregation} module. This rejection criterion preserves the correctness guarantees of speculative decoding while increasing the acceptance probability through structured probability sharing among aligned branches. PathRelax performs parallel verification on all speculative paths in the draft tree. The accepted nodes are then passed to the level-wise selection stage. Among the verified candidates, a longest-common-prefix priority policy is applied: draft nodes exhibiting larger prefix overlap with the most recently accepted token sequence are assigned a higher retention priority and selected as the output of the current iteration.

Tokens preceding the earliest accepted position are discarded, while the remaining unresolved nodes are selectively propagated to the next iteration through a structured tree update procedure. We traverse the draft tree in a predefined level-wise order and preserve only the necessary number of nodes under the capacity constraint imposed at each level, as illustrated in the level-wise selection operation of Fig.~\ref{fig:main}(B). This level-aware reuse strategy preserves the hierarchical organization of the draft tree, controls its growth across iterations, and ensures stable computational complexity. For nodes lacking sufficient historical draft context after pruning, spatial-prior ISP initialization~\cite{SJD} is applied to construct new candidate tokens, thereby restoring the required expansion capacity for subsequent decoding steps. Overall, the relaxed verification mechanism that introduces cross-path probability aggregation improves acceptance efficiency while maintaining the unbiasedness of speculative decoding.

\begin{table*}[t]
\centering
\small
\caption{Performance comparison on the \textbf{MSCOCO2017} validation set. Columns under \textbf{Acceleration} report speedup ratio ($SR$) and mean acceptance length ($\tau$). \textbf{Config} denotes different model inference settings. Temperature is fixed to $T=1$. Abbreviations: LGpt7B (Lumina-gpt-7B-768~\cite{luminamgpt}) and SJD+L (SJD+LANTERN~\cite{LANTERN++}).}
\begin{tabular}{l|l|cc|ccccc}
\toprule
 \textbf{Model}&\textbf{Config} & \multicolumn{2}{c|}{\textbf{Acceleration}} & \textbf{Time (s)$\downarrow$} & \textbf{CLIP$\uparrow$} & \textbf{HPSv2$\uparrow$} & \textbf{FID$\downarrow$} & \textbf{DINO$\uparrow$} \\
  && \textbf{$SR\uparrow$} & \textbf{$\tau\uparrow$} & & & & & \\
\midrule

\multirow{7}{*}{LGpt7B~\cite{luminamgpt}}
 &AR & 1.00× & 1.00 & 106.5& 0.3136 & 0.2904 & 36.25 & 0.1297 \\
 &JD~\cite{JD} & 1.02× & 1.04 & 104.4& 0.3131 & 0.2889 & 35.51 & 0.1288 \\
 &SJD~\cite{SJD} & 1.90× & 2.23 & 56.0& 0.3134 & 0.2905 & 34.01 & 0.1323 \\
 &SJD+L~\cite{LANTERN++} & 2.41× & 3.15 & 44.1& 0.3120 & 0.2845 & 36.52 & 0.1315 \\
 &GSD~\cite{GSD} & 2.58× & 3.31 & 41.2& 0.3121 & 0.2874 & 36.38 & 0.1309 \\
 &SJD++~\cite{SJD++} & 3.12× & \textbf{6.44} & 34.1 & 0.3101& 0.2892& 35.12&0.1247\\
\cmidrule{2-9}
 &\textbf{PathSpec} & \textbf{3.95×} & 5.55 & \textbf{26.9}& 0.3060 & 0.2706 & 38.14 & 0.1288 \\

\midrule

\multirow{7}{*}{Emu3~\cite{EMU3}}
 &AR & 1.00× & 1.00 & 391.3& 0.1904 & 0.2911 & 109.19 & 0.1323 \\
 &JD~\cite{JD} & 0.99× & 1.01 & 392.3& 0.1932 & 0.2928 & 110.29 & 0.1341 \\
 &SJD~\cite{SJD} & 1.76× & 2.32 & 221.4& 0.1904 & 0.2900 & 106.12 & 0.1250 \\
 &SJD+L~\cite{LANTERN++} & 1.75× & 2.31 & 222.5& 0.1897 & 0.2919 & 106.40 & 0.1238 \\
 &GSD~\cite{GSD} & 2.64× & 3.47 & 148.1& 0.1883 & 0.2927 & 112.64 & 0.1204 \\
 &SJD++~\cite{SJD++} & 3.02×& \textbf{6.32} & 129.3& 0.1861& 0.2921& 110.21&0.1243\\
\cmidrule{2-9}
 &\textbf{PathSpec} & \textbf{3.89×} & 5.11 & \textbf{100.6}& 0.1857 & 0.2727 & 113.26 & 0.1163 \\
\bottomrule
\end{tabular}
\label{tab:table2}
\end{table*}

\section{Experiments}
\label{sec:exp}

\noindent\textbf{Datasets:}
For text-conditional image generation, we evaluate the acceleration performance of our method on three widely used benchmarks: Parti-Prompts~\cite{parti-prompts}, MS-COCO 2017~\cite{MSCOCO}, and T2ICompBench~\cite{T2I-CompBench}.
These datasets cover both open-ended prompt generation and compositional reasoning scenarios, providing a comprehensive assessment of decoding efficiency and generation quality.

\noindent\textbf{Acceleration Metrics:}
To quantify inference acceleration, we report the following metrics: \textbf{Speedup Ratio (SR)}: The end-to-end inference speedup relative to vanilla visual autoregressive decoding. \textbf{Acceptance Length ($\tau$)}: The average number of tokens accepted per draft--verification cycle, measuring how effectively the target visual autoregressive model advances in each iteration.

\noindent\textbf{Image Quality Metrics:}
To ensure that acceleration does not degrade generation quality, we evaluate the generated images using standard metrics, including CLIP-score~\cite{clipscore}, HPSv2~\cite{hpsv2}, Inception Score (IS)~\cite{IS}, Fréchet Inception Distance (FID)~\cite{FID}, and DINO score~\cite{DINO1,DINO2}.
For Parti-Prompts, which provide prompts without ground-truth images, we report CLIP-score, HPSv2~\cite{hpsv2}, and IS, and omit FID and DINO scores. For T2ICompBench, we adopt the composition-specific metrics defined in the benchmark: Disentangled BLIP-VQA for attribute binding (color, shape, texture), UniDet-based metrics for spatial relationships (2D/3D) and numeracy, and a composite "3-in-1" metric (averaging CLIPScore~\cite{clipscore}, BLIP-VQA~\cite{BLIP}, and UniDet~\cite{Zhou_2022_CVPR}) for complex compositions.

\noindent\textbf{Implementation Details:}
For both Lumina-GPT-7B-768 and Emu3, we follow standard evaluation practice~\cite{EMU3,luminamgpt} and set the classifier-free guidance scale to 3.0.
In PathSpec, we set the cross-path relaxation coefficient to $\lambda=0.01$.
The sampling temperature is fixed to $T=1$ for all experiments. To evaluate speed, we measure the real latency on 1x NVIDIA GeForce RTX 4090 device. 
\begin{table*}[t]
\setlength{\tabcolsep}{3pt}
\centering
\footnotesize
\caption{The evaluation on the validation set of \textbf{T2ICompBench}. Speedup ratio is denoted by \( SR \), and the mean acceptance length is denoted by \( \tau \). \textbf{Config} denotes different model inference settings. The temperature is \( T=1 \). Abbreviations: LGpt7B (Lumina-gpt-7B-768~\cite{luminamgpt}) and SJD+L (SJD+LANTERN~\cite{LANTERN++}).}
\begin{tabular}{l|l|cc|cccllllc}
\toprule
\textbf{Model}&\textbf{Config} & \multicolumn{2}{c|}{\textbf{Acceleration}} & \textbf{Time (s)$\downarrow$} & \textbf{CLIP$\uparrow$} & \textbf{Color$\uparrow$}&  \textbf{Shape$\uparrow$}& \textbf{ Texture$\uparrow$}&  \textbf{ 2D$\uparrow$}&\textbf{ 3D$\uparrow$}&\textbf{ 3-in-1$\uparrow$}\\
&& \textbf{$SR\uparrow$} & \textbf{$\tau\uparrow$} & & & &  & &  &&\\

\midrule

\multirowcell{7}[0ex][c]{\rotatebox{90}{LGpt7B~\cite{luminamgpt}}}
 &AR & 1.00×& 1.00  & 106.5&0.3246&0.5482&0.3304& 0.4129&  0.2006&0.2837&0.3179\\
 &JD~\cite{JD}& 1.02×& 1.03& 104.4& 0.3213& 0.5474&  0.3325& 0.4218&  0.1954&0.2723&0.3148\\
 &SJD~\cite{SJD} & 1.93×& 2.29  & 55.1& 0.3156 &0.5603&0.3293&0.4109&0.1995&0.2821&0.3165\\
 &SJD+L~\cite{LANTERN++} & 2.18×& 3.05  & 48.7& 0.3154 & 0.5274&  0.3194& 0.4254&  0.1847&0.2251&0.2913\\
 &GSD   & 2.84×& 3.41  & 37.4& 0.3224 & 0.5112&  0.3248& 0.4001&  0.1756&0.2121&0.2726\\
 &SJD++~\cite{SJD++} & 3.81×& \textbf{6.42}& 27.9& 0.3217 &0.5578&0.3279&0.4107&0.1996&0.2799&0.3163\\

\cmidrule{2-12}
 &\textbf{PathSpec} &\textbf{4.18×}&5.62&\textbf{25.4}&       0.3121&       0.5397&   0.3218& 0.4002&  0.1712&0.2132&0.2712\\

\midrule

\multirowcell{7}[0ex][c]{\rotatebox{90}{Emu3~\cite{EMU3}}}
 &AR  &1.00×&       1.00  &       392.1&       0.2246 &0.6549&0.3663&0.5061&0.2066&0.3105&0.3607\\
 &JD~\cite{JD}&       1.01×&       1.03&       388.2&       0.2192&       0.6523&   0.3632& 0.5054&  0.2046&0.3048&0.3647\\
 &SJD~\cite{SJD}   &       1.73×&       2.31 &       226.5&       0.2182 &0.6525&0.3628&0.5046&0.2068&0.3089&0.3595\\
 &SJD+L~\cite{LANTERN++} & 2.33×& 3.08  & 168.1& 0.2104 & 0.6326&  0.3512& 0.4823&  0.1912&0.2929&0.3317\\
 &GSD   &       2.60×&       3.47 &       150.5&       0.2170 &       0.6412&   0.3410& 0.4684&  0.1836&0.3019&0.3285\\
& SJD++~\cite{SJD++} & 3.14×& \textbf{7.48}& 124.5& 0.2189 &0.6508&0.3609&0.5029&0.2071&0.3069&0.3583\\

\cmidrule{2-12}
 &\textbf{PathSpec} &       \textbf{3.82×}&       5.09&       \textbf{102.6}&       0.2196 &       0.6428&   0.3516& 0.4973&  0.1974&0.2934&0.3392\\
\bottomrule
\end{tabular}%
\label{tab:table3}%

\end{table*}%
\begin{table*}[htbp]
\centering
\small
\caption{The evaluation on the validation set of \textbf{Parti-Prompts}. Speedup ratio is denoted by $SR$, and mean acceptance length is denoted by $\tau$. Temperature is fixed to $T=1$. \textbf{Config} denotes different model inference settings. Abbreviations: LGpt7B (Lumina-gpt-7B-768~\cite{luminamgpt}) and SJD+L (SJD+LANTERN~\cite{LANTERN++}).}
\begin{tabular}{l|l|cc|ccccc}
\toprule
\textbf{Model} & \textbf{Config} & \multicolumn{2}{c|}{\textbf{Acceleration}} & \textbf{Time (s)$\downarrow$} & \textbf{CLIP$\uparrow$} & \textbf{HPSv2$\uparrow$} & \textbf{IS$\uparrow$} \\
& & \textbf{$SR\uparrow$} & \textbf{$\tau\uparrow$} & & & & \\
\midrule

\multirow{7}{*}{LGpt7B~\cite{luminamgpt}}
& AR & 1.00× & 1.00 & 106.5 & 0.3225 & 0.2930 & 21.81 \\
& JD~\cite{JD} & 1.01× & 1.04 & 105.4 & 0.3217 & 0.2927 & 21.66 \\
& SJD~\cite{SJD} & 1.88× & 2.29 & 56.5 & 0.3214 & 0.2930 & 21.40 \\
& SJD+L~\cite{LANTERN++} & 2.64× & 3.25 & 40.4 & 0.3192 & 0.2877 & 21.38 \\
& GSD~\cite{GSD} & 2.52× & 3.41 & 42.3 & 0.3189 & 0.2897 & 19.84 \\
& SJD++~\cite{SJD++} & 3.12× & \textbf{6.40} & 34.1 & 0.3217 & 0.2896 & 21.25 \\

\cmidrule{2-8}
& \textbf{PathSpec} & \textbf{4.14×} & 5.64 & \textbf{25.6} & 0.3028 & 0.2627 & 19.23 \\

\midrule

\multirow{7}{*}{Emu3~\cite{EMU3}}
& AR & 1.00× & 1.00 & 390.7 & 0.2281 & 0.2904 & 15.20 \\
& JD~\cite{JD} & 0.99× & 1.02 & 392.7 & 0.2204 & 0.2921 & 15.46 \\
& SJD+L~\cite{SJD} & 1.75× & 2.30 & 224.1 & 0.2214 & 0.2925 & 14.66 \\
& LANTERN~\cite{LANTERN++} & 1.76× & 2.32 & 221.5 & 0.2151 & 0.2923 & 14.55 \\
& GSD~\cite{GSD} & 2.67× & 3.53 & 146.1 & 0.2129 & 0.2927 & 13.96 \\
& SJD++~\cite{SJD++} & 3.01× & \textbf{7.51} & 129.8 & 0.2189 & 0.2914 & 15.11 \\

\cmidrule{2-8}
& \textbf{PathSpec} & \textbf{4.00×} & 5.28 & \textbf{97.6} & 0.2071 & 0.2780 & 13.11 \\

\bottomrule
\end{tabular}
\label{tab:table1}
\end{table*}
\subsection{Main Results}
\label{sec:results}

We comprehensively evaluate the acceleration performance of our method alongside baselines on three benchmarks: Parti-Prompts~\cite{parti-prompts}, T2ICompBench~\cite{T2I-CompBench}, and MSCOCO2017~\cite{MSCOCO} validation sets. 
Accelerated results for MSCOCO2017, T2ICompBench, and Parti-Prompts are presented in Table~\ref{tab:table2}, Table~\ref{tab:table3}, and Table~\ref{tab:table1}, respectively.

PathSpec integrates tree-based parallel exploration with cross-path relaxed validation to jointly optimize SR, $\tau$, and inference time. Results on MSCOCO2017 and T2ICompBench (Tables~\ref{tab:table2}, \ref{tab:table3}) show that it consistently achieves the best acceleration. As shown in Table~\ref{tab:table1}, PathSpec attains top performance on all acceleration metrics, except slightly lower $\tau$ on Parti-Prompts. For Lumina-GPT-7B-768, PathSpec achieves 3.95$\times$--4.18$\times$ speedup with $\tau$ of 5.55--5.64; for Emu3, speedup reaches 3.82$\times$--4.00$\times$ with $\tau$ of 5.09--5.28. In both cases, forward inference time is reduced by over 70\% compared to the baseline. Although minor image quality degradation is observed, it does not significantly affect perceptual fidelity. Overall, PathSpec achieves a strong balance between efficiency and quality. Additional results are provided in Appendix B.2.

\subsection{Ablations and Analysis}

\begin{figure*}[t]
    \centering
    \begin{minipage}[t]{0.3\textwidth}
        \centering
        \includegraphics[width=\textwidth]{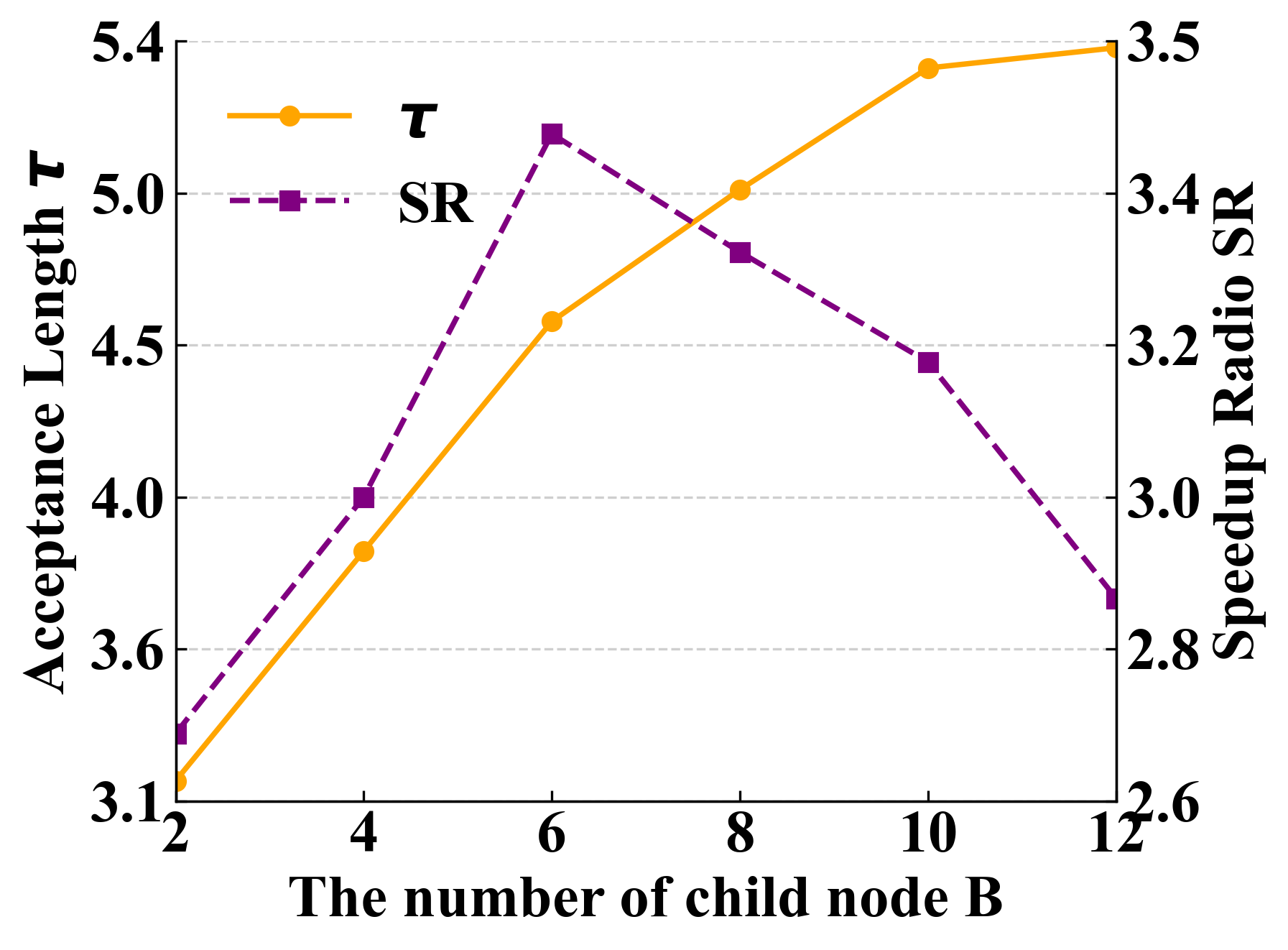}
        \subcaption{}
    \end{minipage}
    \begin{minipage}[t]{0.3\textwidth}
        \centering
        \includegraphics[width=\textwidth]{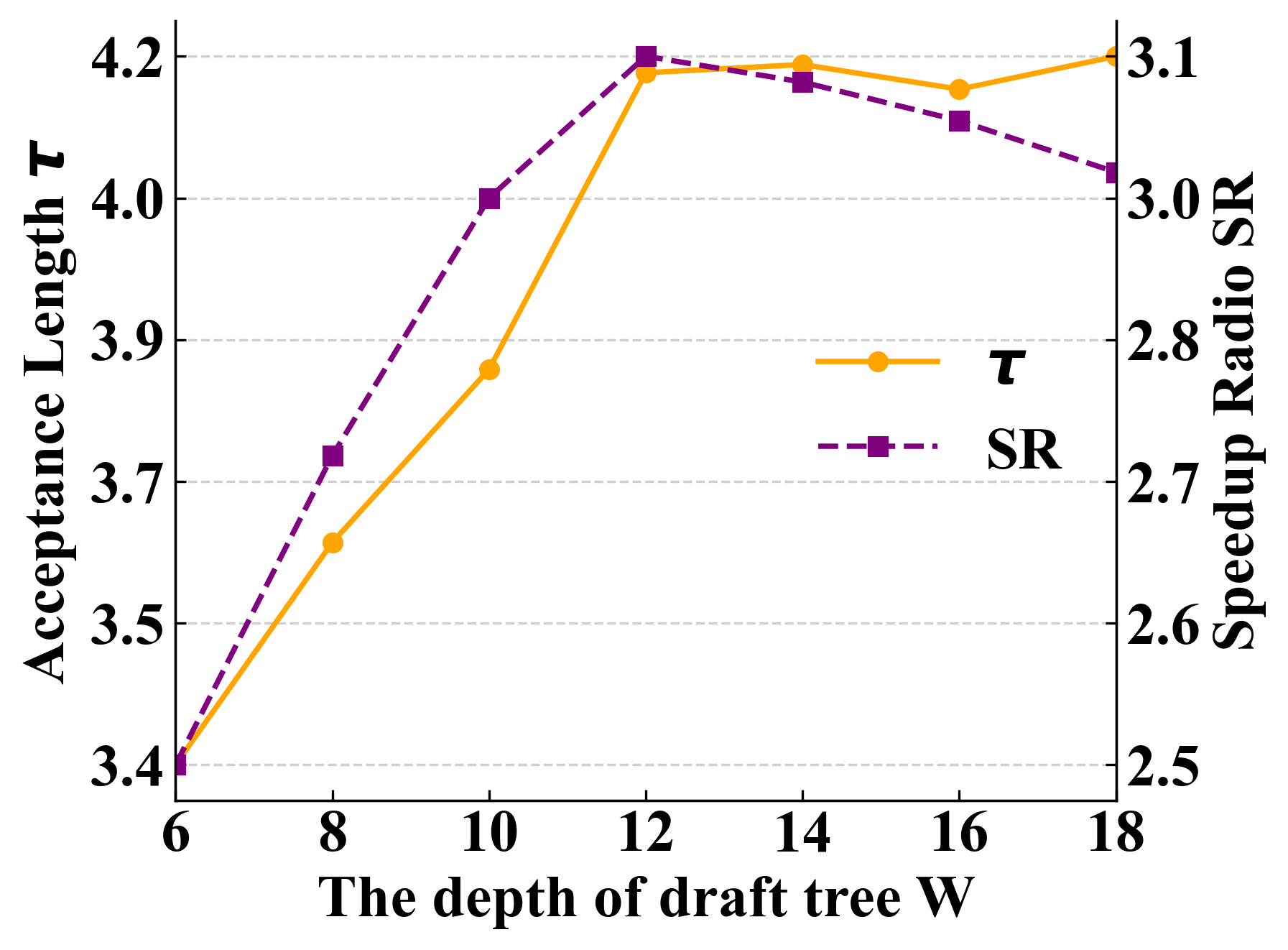}
        \subcaption{}
    \end{minipage}
    \begin{minipage}[t]{0.3\textwidth}
        \centering
        \includegraphics[width=\textwidth]{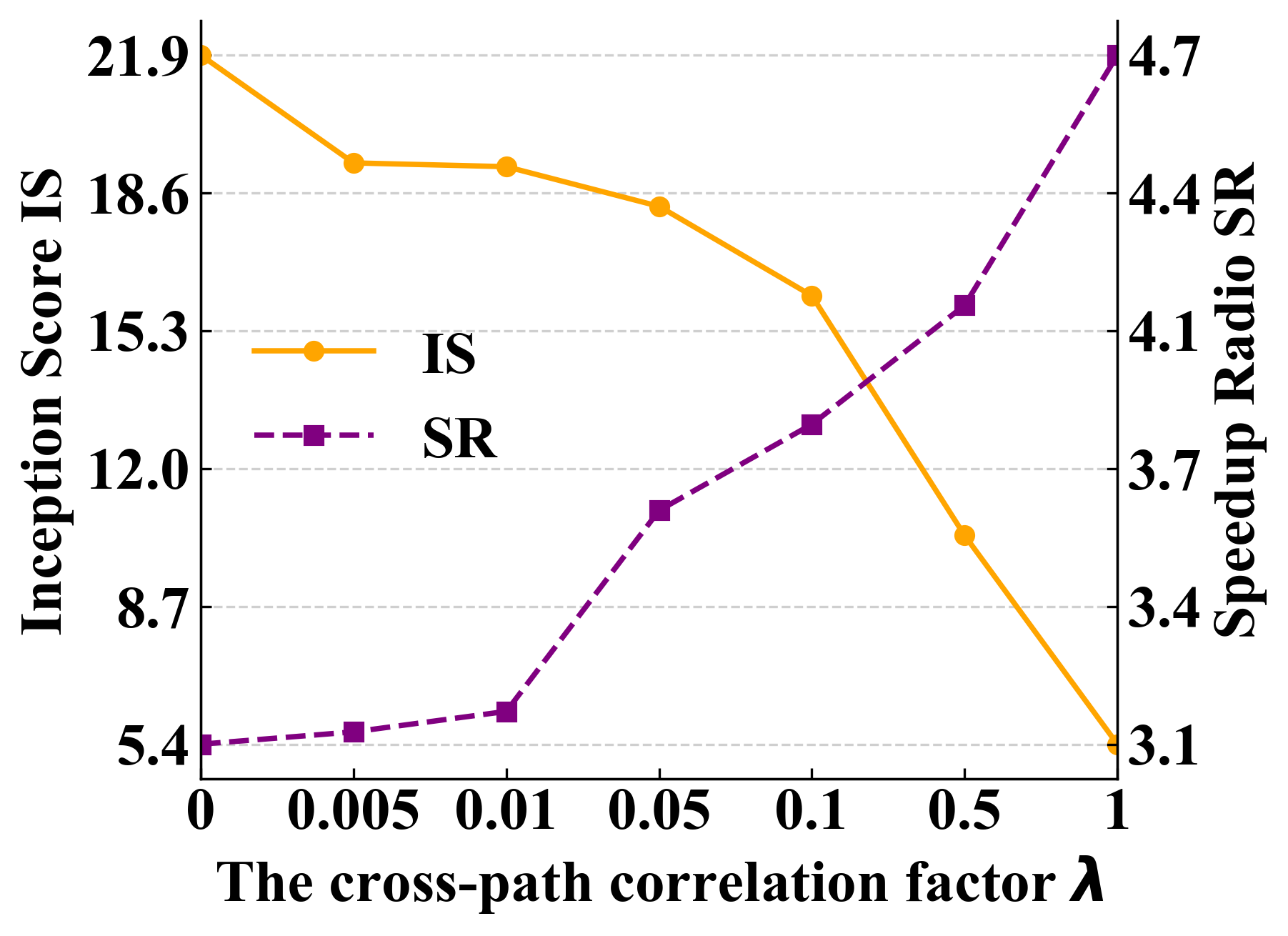}
        \subcaption{}
        \label{fig:d_IS_SR}
    \end{minipage}
    \caption{\textbf{Comparison of acceleration under different hyperparameters}. (a) The relationships of $\tau$ and $SR$ with the maximum number of nodes per level $B$, respectively. (b) $\tau$ and $SR$ with the depth $W$, respectively. (c) The relationships of $IS$ and $SR$ with the number of draft token sequences $\lambda$, respectively. The model used for testing is Lumina-GPT-7B-768~\cite{luminamgpt}.} 
    \label{fig:4.3.1.2}
\end{figure*}

\begin{figure}[t]
    \centering
    \includegraphics[width=0.85\linewidth]{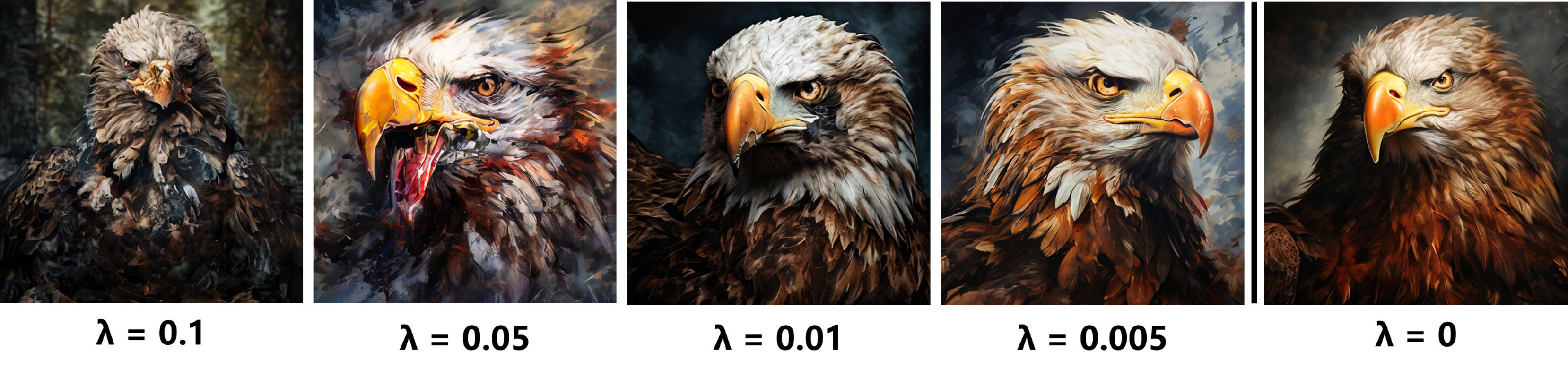}
    \caption{\textbf{The visualization of image quality.} For the prompt "an eagle", generated images across different settings with $\lambda = 0.01$ are provided.}
    \label{fig:IS_λ}
\end{figure}

\noindent\textbf{Selection of the Tree Structure:}
In PathSpec, nodes at the same depth share identical expected acceptance rates, so the draft tree is characterized by depth $W$ and branching factor $B$, with a total node limit of 72. As $B$ increases, the acceptance length $\tau$ also increases; however, beyond a threshold, verification overhead offsets the benefit. For example, although $W=8$ achieves a higher $\tau$ than $W=6$, the added cost—equivalent to two extra branches—reduces the speculative ratio (SR). Overall, $W=6$ is selected as the optimal setting for Lumina-GPT-7B-768 and Emu3.

As shown in Fig.~\ref{fig:4.3.1.2}(b), we initially expected that increasing the depth $W$ of the draft tree would lead to a corresponding improvement in the acceptance length $\tau$. However, experimental results indicate that $\tau$ saturates once $W$ exceeds a certain threshold. Specifically, when $H>12$, further increasing the tree depth results in a substantial growth in the number of draft tokens, while bringing no appreciable gains in $\tau$, which reduces the speedup ratio due to the additional computational overhead.

\noindent\textbf{Selection of $\lambda$ in :}
As shown in Fig.~\ref{fig:4.3.1.2}(c), increasing $\lambda$ relaxes constraints, improving acceleration (SR) but degrading image quality (IS). Although $\lambda=0.05$ achieves a high SR of 3.61, as illustrated in Fig.~\ref{fig:IS_λ}, it leads to notable quality deterioration. To maintain stable image quality, we choose $\lambda=0.01$ for both Lumina and Emu3.

\section{Conclusion}
\label{sec:conclusion}
We propose \textbf{Parallel-Path Cross-Relaxed Speculative Jacobi Decoding}, a novel framework that accelerates autoregressive text-to-image generation by extending Speculative Jacobi Decoding with multiple draft sequences and cross-relaxed sampling. We compare the increase in draft token scale from SJD to PathExplore and analyze the relaxed sampling mechanism of PathRelax designed based on the semantic similarity of draft tokens across sequences. Our method achieves speedup ratios of 4.14×, 4.18×, and 3.95× on Parti-Prompts, T2ICompBench, and MSCOCO2017, offering an efficient solution for real-time image generation with broad application potential.

    
{
    \small
    \bibliographystyle{ieeenat_fullname}
    \bibliography{main}
}

\clearpage
\appendix
\begin{figure*}[t]
    \centering
    \begin{minipage}[t]{0.45\textwidth}
        \centering
        \includegraphics[width=\textwidth]{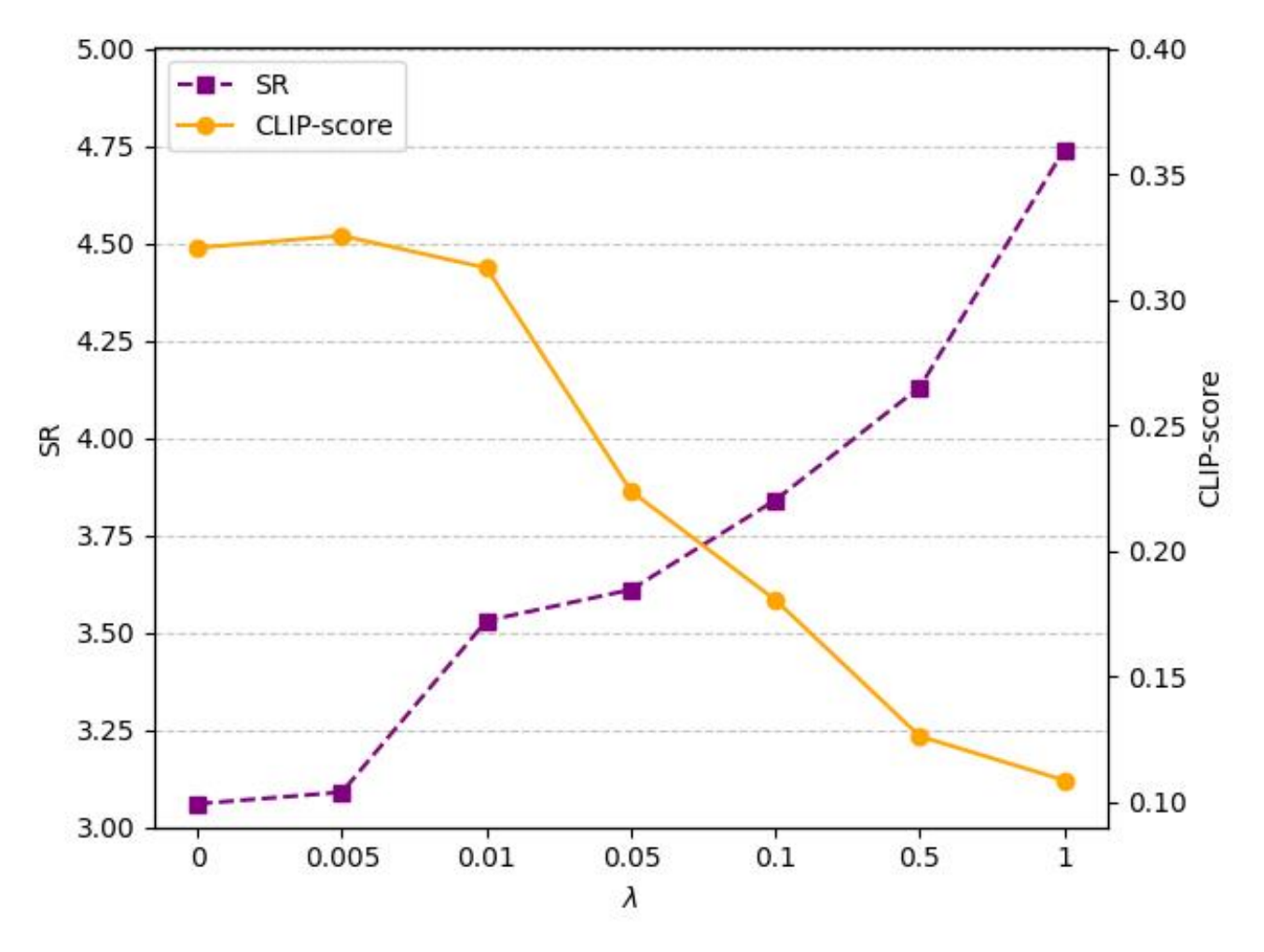}
        \subcaption{}
    \end{minipage}
    \begin{minipage}[t]{0.45\textwidth}
        \centering
        \includegraphics[width=\textwidth]{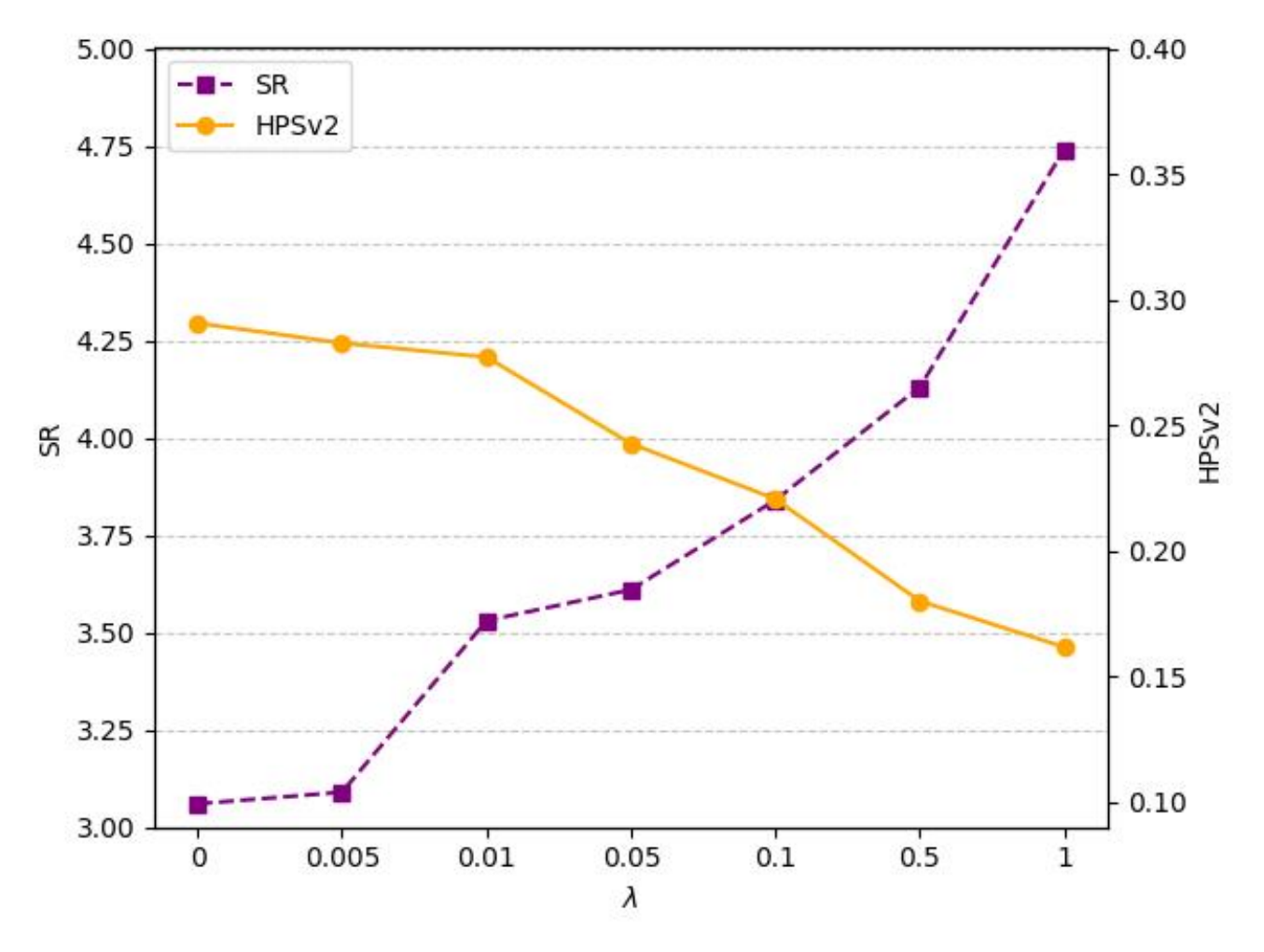}
        \subcaption{}
    \end{minipage}
    \caption{\textbf{Comparison of acceleration and image quality under different hyperparameters}. (a) The relationships of $CLIP-score$~\cite{clipscore} and $SR$ with the number of draft token sequences $\lambda$, respectively. (b) The relationships of $HPSv2$~\cite{hpsv2} and $SR$ with the number of draft token sequences $\lambda$, respectively. The model used for testing is Lumina-GPT-7B-768~\cite{luminamgpt}.} 
    \label{fig:supp1_2}
\end{figure*}

\begin{table*}[htbp]

  \centering
  \caption{\textbf{The performance of different relaxation methods on the validation set of Parti-Prompts.} Speedup ratio is denoted by \( SR \), and the mean acceptance length is denoted by \( \tau \). The temperature is \( T=1 \).}
    \begin{tabular}{l|rr|rrrr}
    \toprule
    \textbf{Configuration} & \multicolumn{2}{c|}{\textbf{Acceleration}} & \multicolumn{1}{l}{\textbf{Time (s)↓}} & \multicolumn{1}{l}{\textbf{CLIP-score↑}} & \multicolumn{1}{l}{\textbf{HPSv2↑}} & \multicolumn{1}{l}{\textbf{IS↑}} \\
          & \multicolumn{1}{l}{\textbf{$SR$↑}} & \multicolumn{1}{l|}{\textbf{$\tau$↑}} &       &       &       &  \\
    \midrule
    Lumina-gpt-7B-768~\cite{luminamgpt} & 1.00×& 1.00  & 106.50 &       0.3225&       0.2930&  21.81\\
    PathExplore & 3.06×& 4.20& 34.60& 0.3206& 0.2906& 21.87\\
    PathExplore+LANTERN& 3.62×& 5.04& 29.40& 0.3166& 0.2826& 21.08\\
    PathExplore+GSD& 3.68×& 5.12& 28.66& 0.3103& 0.2722& 19.62\\
    \midrule
    PathSpec+SJD($\lambda=0.01$)& 3.53×& 4.44& 30.10&       0.3126&       0.2771&  20.68\\
    PathSpec+SJD($\lambda=0.08$)& 3.70×& 5.18& 28.78& 0.3112& 0.2784& 20.86\\
    PathSpec & \textbf{4.14×}& \textbf{5.64}  & \textbf{25.69} & 0.3028 & 0.2627 & 19.23\\
    \bottomrule
    \end{tabular}%
  \label{tab:tablesup1}%

\end{table*}%


\section{Proof of the Lossless Guarantees of PathExplore}
\label{sec:proof_pathexplore}

\noindent
\textbf{Theorem 1} 
The token sequence accepted by the Parallel-Path Speculative Jacobi Decoding (PathExplore) satisfies the target distribution $p_\theta( \mathbf{x} | \mathbf{x}_{1:i-1}^{(j)})$, provided that the cross-path relaxation is disabled (i.e., $\lambda=0$). Here, $\mathbf{x}$ denotes a token, $j$ denotes the iteration index, $i$ denotes the token index, and $\theta$ denotes the autoregressive model parameters.

\vspace{1em}
\noindent
\textit{Proof.} 
PathExplore extends the standard Speculative Jacobi Decoding (SJD) by maintaining a draft token tree $\mathcal{T}^{(j)}$ instead of a single draft chain. However, the verification mechanism for any individual candidate token $\mathbf{x}$ within the tree remains consistent with the rejection sampling principle used in standard speculative decoding.

To prove the correctness of PathExplore, we consider a generic draft token $\mathbf{x}$ generated at a specific node in the draft tree. We demonstrate that the marginal probability of this token being output (either by acceptance or by resampling upon rejection) equals the target probability $p_\theta( \mathbf{x} | \mathbf{x}_{1:i-1}^{(j)})$. Let $\mathcal{J}^{(j)}$ denote the condition with draft tree at the current $j$-th iteration (including the current model weights $\theta$ and the prefix context), and $\mathcal{J}^{(j-1)}$ denote the condition with draft tree from the previous iteration used to generate the draft.
We denote the target probability as $p(\mathbf{x} | \mathcal{J}^{(j)})$ and the draft probability as $p(\mathbf{x} | \mathcal{J}^{(j-1)})$. Note that in PathExplore, the draft probability comes from the specific branch of the tree inherited from the previous iteration.

The process for a node in PathExplore consists of two mutually exclusive outcomes for ensuring the target distribution:
(a) \textbf{Acceptance:} The draft token is accepted based on the verification criterion.
(b) \textbf{Rejection and Resampling:} The draft token is rejected, and a new token is resampled from the modified residual distribution.

\paragraph{1. Acceptance Probability}
According to the PathExplore method (and standard speculative decoding), a candidate token $\mathbf{x}$ from the draft tree is accepted with probability:
\begin{align}
    p(r \text{ is true} | \mathbf{x} , \mathcal{J}^{(j)} , \mathcal{J}^{(j-1)} ) = \min \left\{ 1,  \frac{ p(\mathbf{x} | \mathcal{J}^{(j)}) }{ p(\mathbf{x} | \mathcal{J}^{(j-1)}) } \right\},
\label{eq:pe-ac-prob}
\end{align}
where $r$ is the boolean variable representing acceptance. The joint probability of a token $\mathbf{x}$ being sampled by the draft strategy and subsequently accepted is:
\begin{equation}
\begin{aligned}
    &p(r \text{ is true} ,\mathbf{x}| \mathcal{J}^{(j)} , \mathcal{J}^{(j-1)}) \\=&~ p(\mathbf{x}| \mathcal{J}^{(j-1)}) \cdot p(r \text{ is true} |\mathbf{x}, \mathcal{J}^{(j)} , \mathcal{J}^{(j-1)}) \\
    =&~ p(\mathbf{x}| \mathcal{J}^{(j-1)}) \cdot \min \left\{ 1,  \frac{ p(\mathbf{x}| \mathcal{J}^{(j)}) }{ p(\mathbf{x}| \mathcal{J}^{(j-1)}) } \right\} \\
    =&~ \min \{   p(\mathbf{x}| \mathcal{J}^{(j)}) ,  p(\mathbf{x}| \mathcal{J}^{(j-1)})  \}.
\label{eq:pe-joint-ac}
\end{aligned}
\end{equation}

\paragraph{2. Rejection and Resampling Probability}
If the token is rejected, we must account for the probability mass that was not covered by the acceptance step. The probability of rejection for the draft distribution is:
\begin{equation}
\begin{aligned}
    &p(r \text{ is false} | \mathcal{J}^{(j)} , \mathcal{J}^{(j-1)}) \\
    =&~ 1 - \sum_{x'} p(r \text{ is true}, x' | \mathcal{J}^{(j)} , \mathcal{J}^{(j-1)} ) \\
    =&~ \sum_{x'} p(x' | \mathcal{J}^{(j)}) - \sum_{x'} \min \{   p(x' | \mathcal{J}^{(j)}) ,  p(x' | \mathcal{J}^{(j-1)})  \} \\
    =&~ \sum_{x'} \max \{ 0, p(x' | \mathcal{J}^{(j)}) - p(x' | \mathcal{J}^{(j-1)}) \}.
\label{eq:pe-demon-rej}
\end{aligned}
\end{equation}
To ensure the target distribution is recovered, the resampling probability subsequent to rejection must be calibrated as:
\begin{equation}
\begin{aligned}
    & p(\mathbf{x}| r \text{ is false} ,  \mathcal{J}^{(j)} , \mathcal{J}^{(j-1)} )
    \\ &=
    \frac {\max \{ 0, p(\mathbf{x}| \mathcal{J}^{(j)} ) - p(\mathbf{x}| \mathcal{J}^{(j-1)} ) \} }{ \sum_{x'} \max \{ 0, p(x' | \mathcal{J}^{(j)} ) - p(x' | \mathcal{J}^{(j-1)} )  \} }.
\label{eq:pe-resample-prob}
\end{aligned}
\end{equation}
Combining Eq.~\ref{eq:pe-demon-rej} and Eq.~\ref{eq:pe-resample-prob}, the joint probability of generating $\mathbf{x}$ via the rejection-resampling path is:
\begin{equation}
\begin{aligned}
    &p(\mathbf{x}| r \text{ is false} , \mathcal{J}^{(j)} , \mathcal{J}^{(j-1)} ) \cdot p(r \text{ is false} | \mathcal{J}^{(j)} , \mathcal{J}^{(j-1)} ) \\
    =&~ \max \{ 0, p(\mathbf{x}| \mathcal{J}^{(j)} ) - p(\mathbf{x}| \mathcal{J}^{(j-1)} ) \}.
\label{eq:pe-joint-rej}
\end{aligned}
\end{equation}

\paragraph{3. Total Probability}
We verify that the sum of probabilities from both cases recovers the target distribution $p(\mathbf{x}| \mathcal{J}^{(j)})$. Using the identity $a = \min(a, b) + \max(0, a-b)$, we have:
\begin{equation}
\begin{aligned}
    &p(\mathbf{x}| \mathcal{J}^{(j)}) \\
    =&~ \min \{   p(\mathbf{x}| \mathcal{J}^{(j)}) ,  p(\mathbf{x}| \mathcal{J}^{(j-1)})  \} + \max \{ 0, p(\mathbf{x}| \mathcal{J}^{(j)} ) - p(\mathbf{x}| \mathcal{J}^{(j-1)} ) \} \\
    =&~ p(r \text{ is true} ,\mathbf{x}| \mathcal{J}^{(j)} , \mathcal{J}^{(j-1)}) + p(r \text{ is false} ,\mathbf{x}| \mathcal{J}^{(j)} , \mathcal{J}^{(j-1)}).
\label{eq:pe-proof-done}
\end{aligned}
\end{equation}
According to Eq.~\ref{eq:pe-proof-done}, the conditional distribution of any token processed by the verification step exactly matches $p(\mathbf{x}| \mathcal{J}^{(j)})$.

\paragraph{Conclusion regarding Parallel Paths}
PathExplore generates multiple candidate branches in parallel. Let $\mathcal{P}_k$ be the $k$-th path in the draft tree. Since the verification of every node in $\mathcal{P}_k$ strictly follows the rejection sampling criterion described above, every validated token in the tree is an unbiased sample from $p_\theta$. Specifically, assuming the root node is $\mathbf{x}_{L}$, the aforementioned condition associated with the draft tree can be formulated as $\mathcal{J}^{(j)} = \{\mathbf{x}^{(j)}_{1:L}, \mathcal{P}_k\}$. PathExplore selects the path with the longest accepted prefix; since every token within that prefix has satisfied the acceptance condition $r$ (or was resampled from the correct residual), the entire resulting sequence $\mathbf{x}_{1:L}^{(j)}$ is guaranteed to follow the target distribution $p_\theta( \mathbf{x} |\mathbf{x}_{1:i-1}^{(j)})$. Thus, PathExplore is lossless.

\begin{figure*}[t!]
    \centering
    \includegraphics[width=1\linewidth]{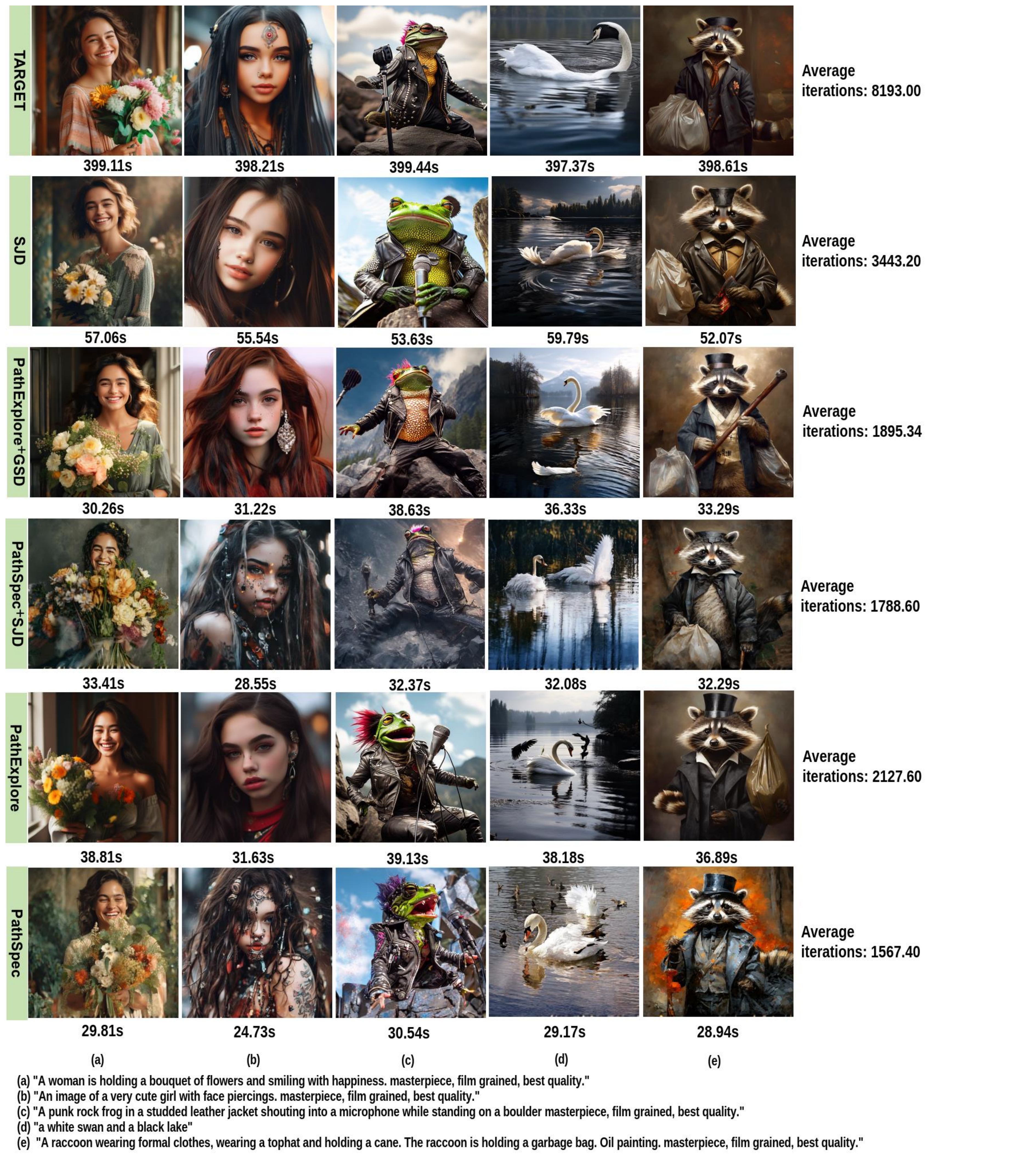}
    \caption{\textbf{The visualization of image quality.} The model used in this experiment is Emu3~\cite{EMU3}. Below each generated image, the time required to generate that image is displayed (excluding the decoder decoding time). On the far right of the images, the number of iteration steps of the transformer architecture autoregressive (AR) model is shown.}
    \label{fig:supvisual}
\end{figure*}
\section{More Experiments}
\subsection{More Experiments of the selection of $\lambda$}

To further elucidate the influence of the regularization coefficient $\lambda$ on final image quality, we conduct a more comprehensive evaluation using multiple metrics. As illustrated in Fig.~\ref{fig:supp1_2}, beyond the Inception Score (IS) already reported above, we additionally include Human Preference Score v2 (HPSv2) and CLIP-Score for assessment. It can be observed that when $\lambda = 0.05$, both image quality metrics exhibit a noticeable decline. Taking both generative quality and speedup ratio into joint consideration, we ultimately retain $\lambda = 0.01$ as our chosen value.

\subsection{Ablation Experiments about Relaxed Sampling}

Table~\ref{tab:tablesup1} presents the performance of different relaxation methods on the Parti-Prompts dataset, where PathExplore+LANTERN uses $\delta=3$, and GSD uses 7 clusters, corresponding to the official settings of LANTERN++~\cite{LANTERN++} and GSD~\cite{GSD}, respectively. Although PathExplore achieves similar acceleration performance compared to PathSpec+SJD ($\lambda=0.08$), its image quality is noticeably degraded.

Moreover, Fig.~\ref{fig:supvisual} shows the visual generation results of different methods. The item of TARGET refers to the quality achieved by the vanilla method in Emu3. As shown, it requires nearly 400 seconds for generation, with an average of 8193 iterations. The SJD method effectively reduces the generation time to approximately 60 seconds while lowering the average number of iterations to around 3500. The further improved PathExplore reduces the time to about 40 seconds, with the average iterations dropping to 2127. Both SJD and PathExplore are lossless with respect to image quality, which has already been theoretically proven in the original paper of speculative sampling~\cite{SSM, SSM1}.

When combined with additional relaxation techniques such as GSD, the generation time is further reduced to approximately 35 seconds, and the average iterations decrease to 1895. PathSpec+SJD alone brings the generation time down to about 32 seconds, with an average of 1788 iterations. The combination of PathRelax and GSD (i.e., PathSpec) achieves the best speedup, reducing the generation time to approximately 28 seconds and the average iterations to 1567.

Therefore, in real-world applications requiring high-fidelity generation—such as detailed human portraits or other fine-grained tasks—PathExplore is recommended. It delivers acceleration superior to that of existing relaxation methods like GSD and LANTERN++, while preserving or even improving image quality. In contrast, when generation speed is the primary concern—such as for coarse-grained tasks like landscape synthesis—PathSpec+SJD or PathSpec is the preferred choice.

\subsection{Ablation Study on PathExplore and PathRelax Modules}
\label{sec:Ablations and Analysis}
To evaluate the effectiveness of PathExplore and PathRelax, we conduct ablation studies comparing them with SJD. As shown in Table~\ref{tab:Ablations_on_MSCR}, PathExplore is a lossless method that removes all relaxed sampling constraints while preserving full speculative decoding fidelity. It achieves substantial improvements over the SJD baseline, yielding 83\% and 64\% increases in average acceptance length on Lumina-GPT-7B-768 and Emu3, respectively, while reducing inference time by 39\%. Furthermore, PathRelax, which replaces GSD with relaxed sampling, delivers even higher acceptance length gains of 94\% and 92\% on the same models, accompanied by 47\% and 48\% reductions in inference time. These results demonstrate that PathExplore provides greater overall gains than PathRelax. Optimizing the structure of draft token paths yields more substantial benefits than merely relaxing acceptance criteria, while maintaining lossless generation quality.

\section{Acceptance Probability of a Draft Token}

We can compute the acceptance probability for a given position within the PathExplore framework. The acceptance rate of a draft distribution in the SJD framework can be expressed as:
\begin{equation}
\begin{split}
\alpha^k_i 
&= 1 - \frac{1}{2} \sum_{x\in \mathcal{V}} \left[ p^k(\mathbf{x}\mid x_{1:i-1}) - q^k(\mathbf{x}\mid x_{1:i-1}) \right] \\
&= 1 - \frac{1}{2} \sum_{x\in \mathcal{V}} \left[ p^{k}_\theta(\mathbf{x}\mid x^j_{1:i-1}) - p^{k}_\theta(\mathbf{x}\mid x^{j-1}_{1:i-1}) \right] \\
&=1-TV(p^{k}_\theta( \cdot \mid x^j_{1:i-1}), p^{k}_\theta( \cdot \mid x^{j-1}_{1:i-1})),
\end{split}
\end{equation}
where $p_\theta^k(\mathbf{x}\mid x_{1:i-1}^j)$ denotes the conditional probability of token $x$ at position $i$ in the $k$-th sequence, computed during the $j$-th forward pass of the target model. 

\begin{table*}[t]
  \centering
  \small
  \caption{\textbf{Performance comparison with different module configurations.} LGpt7B is Lumina-gpt-7B-768~\cite{luminamgpt}.}
    \begin{tabular}{l|l|cc|c}
    \toprule
    \textbf{Model}& \multicolumn{1}{c|}{\textbf{Configuration}} & \multicolumn{2}{c|}{\textbf{Acceleration}} & \multicolumn{1}{l}{\textbf{Time (s)$\downarrow$}} \\
    & & \textbf{$SR\uparrow$} & \textbf{$\tau\uparrow$} & \\
    \midrule
    LGpt7B~\cite{luminamgpt}& SJD   & 1.88× & 2.29  & 56.5 \\
          & \textbf{PathExplore} & 3.06× & 4.20(+83\%)& 34.6(-39\%)\\
          & \textbf{PathRelax} & 3.53× & 4.44(+94\%)& 30.1(-47\%)\\
    \midrule
    Emu3~\cite{EMU3}& SJD   & 1.88× & 2.30   & 224.1 \\
          & \textbf{PathExplore} & 2.86× & 3.77(+64\%)& 136.5(-39\%)\\
          & \textbf{PathRelax} & 3.35× & 4.42(+92\%)& 116.6(-48\%)\\
    \bottomrule
    \end{tabular}%
  \label{tab:Ablations_on_MSCR}%
\end{table*}%
\section{Substitutability of the Image Tokens}

\begin{figure}[t]
	\centering
	\begin{minipage}[t]{0.45\textwidth}
		\centering
		\includegraphics[width=\textwidth]{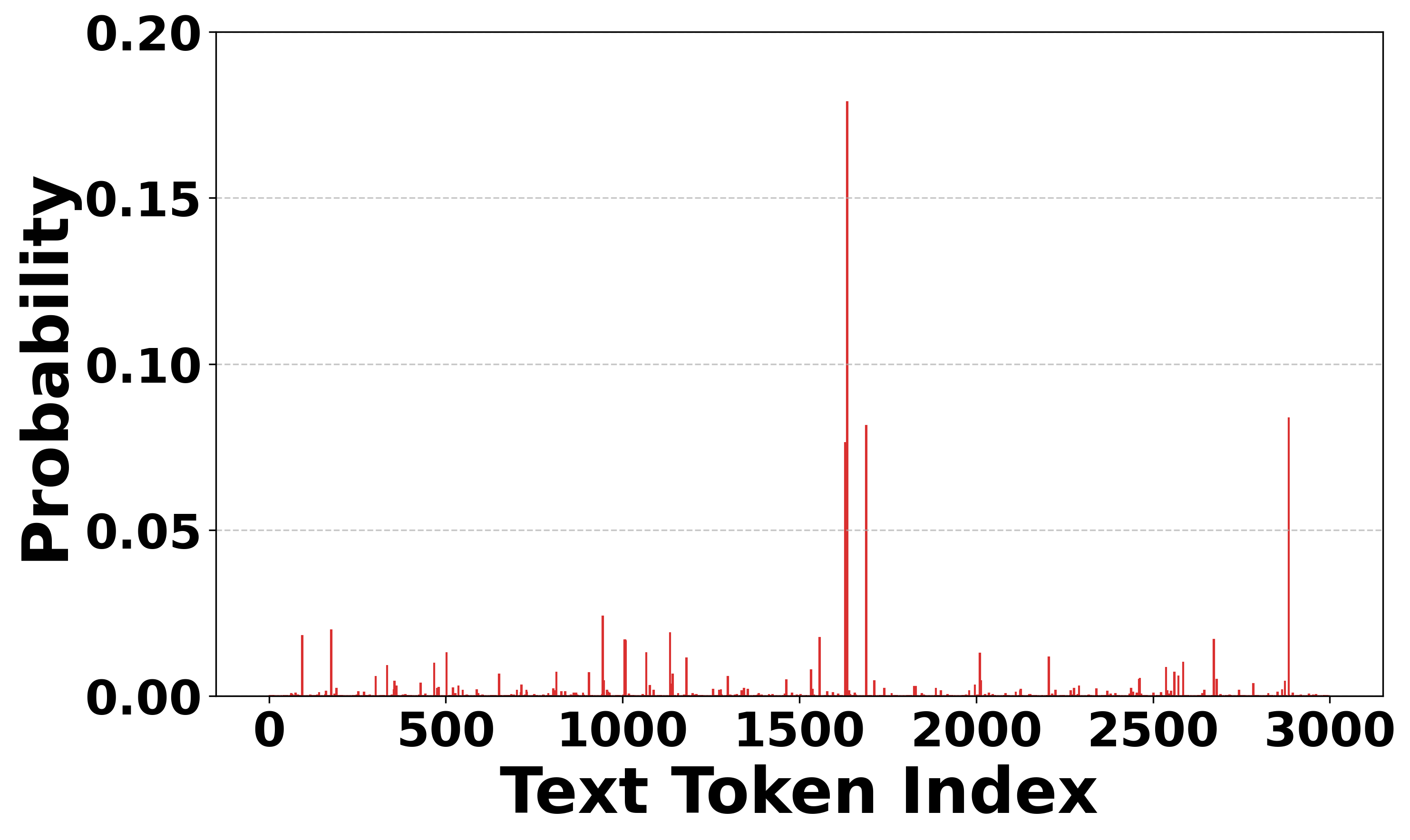}
		\subcaption{}
	\end{minipage}
	\begin{minipage}[t]{0.45\textwidth}
		\centering
		\includegraphics[width=\textwidth]{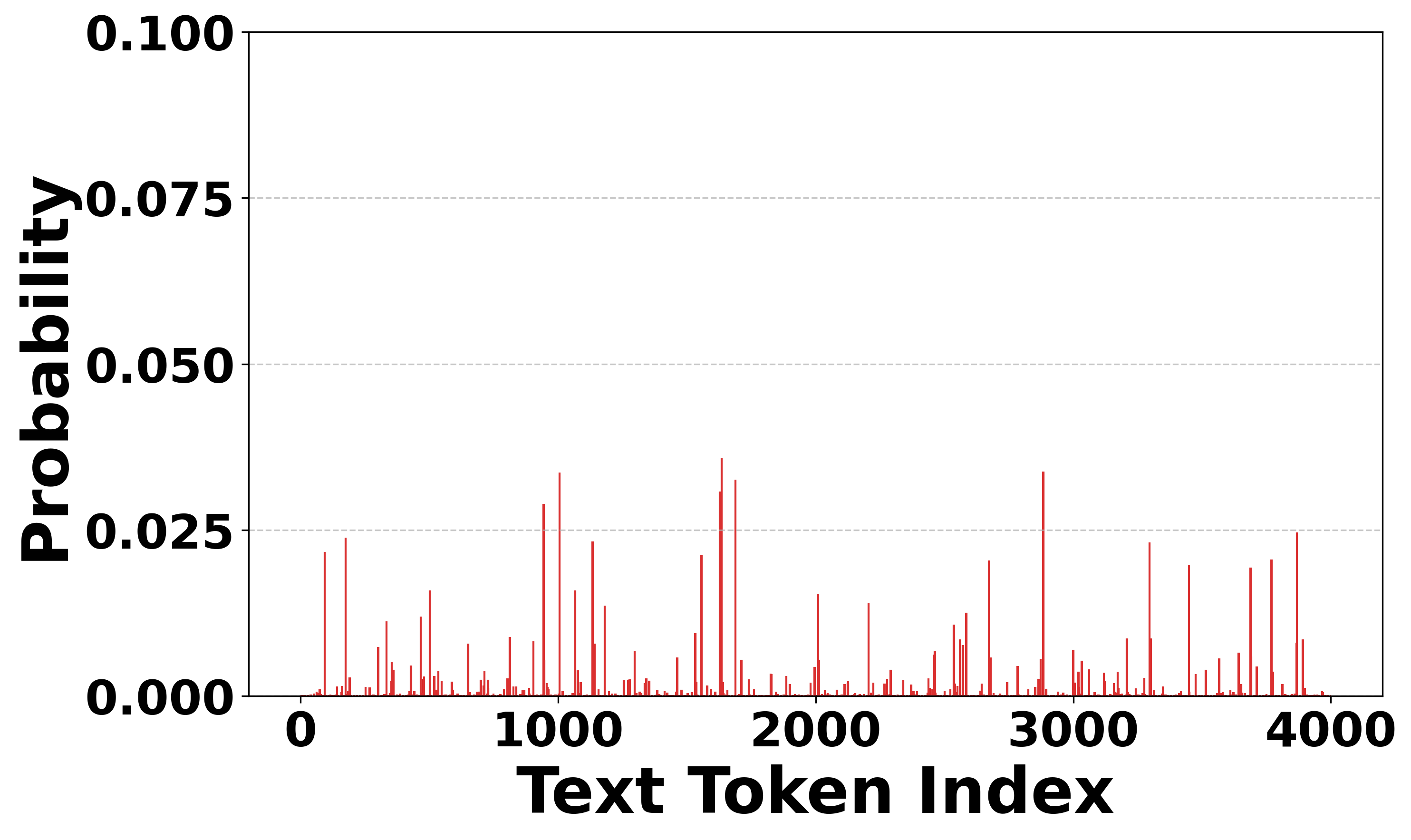}
		\subcaption{}
	\end{minipage}
	\caption{\textbf{The differences between image token and text token generation to show image token at the same position share similar probability and have some freedom to sample.} (a) The probability distribution chart of text token generation at a certain moment. (b) The probability distribution chart of image token generation at a certain moment. By comparing (a) and (b), it can be observed that the generation distribution of image tokens exhibits greater ambiguity. 
    }
    \label{fig:text_image_p}
    \vspace{-1em}   
\end{figure}

As illustrated in Fig.~\ref{fig:text_image_p}(a), it shows the next-token probability distribution of Vicuna-13B, where the peak probability exceeds 0.15 and many tokens have probabilities near zero. The predictive distributions of LLMs are characterized by a few dominant tokens with high probabilities. In contrast, AR visual models do not exhibit such peakedness. As shown in Fig.~\ref{fig:text_image_p}(b), the distribution for Lumina-GPT is notably flatter; its maximum probability remains below 0.05, with numerous tokens exhibiting probabilities comparable to the peak. These observations suggest that text generation models are inherently more deterministic. 
Conversely, during autoregressive text-to-image generation, image token distributions are frequently characterized by high ambiguity, with multiple tokens exhibiting similar semantics and probabilities. This implies that the sampled image tokens exhibit a certain degree of replaceability within a range. Prior relaxed sampling methods (e.g., GSD~\cite{GSD}, LANTERN~\cite{LANTERN}) exploit this ambiguity within a single sequence, relaxing verification against the codebook. However, their single-sequence designs restrict relaxation to one token at one position, leading to strong dependence on codebook quality and heuristic hyperparameter tuning.

\end{document}